\documentclass{article}

\usepackage{microtype}
\usepackage{graphicx}
\usepackage{subfigure}
\usepackage{booktabs} 

\usepackage{hyperref}

\usepackage{varwidth}
\usepackage{booktabs}
\usepackage{arydshln}
\usepackage{amsmath}
\usepackage{amsthm}
\usepackage{amssymb}
\newcommand{\R}{\mathbb{R}}
\newtheorem{theorem}{Theorem}
\newtheorem{corollary}{Corollary}
\usepackage{longtable}
\usepackage[accepted]{icml2018}

\icmltitlerunning{Open Category Detection with PAC Guarantees}

\begin{document}

\twocolumn[
\icmltitle{Open Category Detection with PAC Guarantees}



\icmlsetsymbol{equal}{*}

\begin{icmlauthorlist}
\icmlauthor{Si Liu}{equal,osus}
\icmlauthor{Risheek Garrepalli}{equal,osu}
\icmlauthor{Thomas G.~Dietterich}{osu}
\icmlauthor{Alan Fern}{osu}
\icmlauthor{Dan Hendrycks}{uc}
\end{icmlauthorlist}

\icmlaffiliation{osus}{Department of Statistics, Oregon State University, Oregon, USA}
\icmlaffiliation{osu}{School of EECS, Oregon State University, Oregon, USA}
\icmlaffiliation{uc}{University of California, Berkeley, California, USA}

\icmlcorrespondingauthor{Si Liu}{lius2@oregonstate.edu}

\icmlkeywords{Anomaly Detection, Open Category Detection, Robust AI}

\vskip 0.3in
]



\printAffiliationsAndNotice{\icmlEqualContribution} 

\begin{abstract}
Open category detection is the problem of detecting ``alien" test instances that belong to categories or classes that were not present in the training data. In many applications, reliably detecting such aliens is central to ensuring the safety and accuracy of test set predictions. Unfortunately, there are no algorithms that provide theoretical guarantees on their ability to detect aliens under general assumptions. Further, while there are algorithms for open category detection, there are few empirical results that directly report alien detection rates. Thus, there are significant theoretical and empirical gaps in our understanding of open category detection. In this paper, we take a step toward addressing this gap by studying a simple, but practically-relevant variant of open category detection. In our setting, we are provided with a ``clean" training set that contains only the target categories of interest and an unlabeled ``contaminated'' training set that contains a fraction $\alpha$ of alien examples. Under the assumption that we know an upper bound on $\alpha$, we develop an algorithm with PAC-style guarantees on the alien detection rate, while aiming to minimize false alarms. Empirical results on synthetic and standard benchmark datasets demonstrate the regimes in which the algorithm can be effective and provide a baseline for further advancements. 
\end{abstract}

\section{Introduction}
\label{submission}

Most machine learning systems implicitly or explicitly assume that their training experience is representative of their test experience. This assumption is rarely true in real-world deployments of machine learning, where ``unknown unknowns", or ``alien" data, can arise without warning. Ignoring the potential for such aliens can lead to serious safety concerns in many applications and significantly degrade the accuracy of test set predictions in others. For example, consider a scientific application where a classifier is trained to recognize specific categories of insects in freshwater samples in order to detect important environmental changes \cite{lytle2010}. Test samples will typically contain some fraction of specimens belonging to species not represented in the training data. A classifier that is unaware of these new species will misclassify the specimens as belonging to existing species. This will produce incorrect scientific conclusions. 

The problem of open category detection is to detect such alien examples at test time. An ideal algorithm for this problem would guarantee a user-specified alien-detection rate (e.g., 95\%), while attempting to minimize the false alarm rate. Unfortunately, no existing algorithm provides such guarantees under general conditions. In addition, empirical evaluations of existing algorithms for open category detection typically do not directly evaluate alien detection rates, which are perhaps the most relevant for safety-critical applications. Overall, our current theoretical and practical understanding of open category detection is lacking from a safety and accuracy perspective. 

\emph{Is it possible to achieve open category detection with guarantees?} In this paper, we take a step toward answering this question by studying a simplified, but practically relevant, problem setting. To motivate our setting, consider the above insect identification problem. At training time it is reasonable to expect that a clean training set is available that contains only the insect categories of interest. At test time, a new sample will include insects from the training categories along with some percentage of insects from new alien categories. Further, scientists may have reasonable estimates for this percentage based on their scientific knowledge and practical experience. We would like to guarantee that the system is able to raise an alarm for, say, 95\% of the insects from alien classes, with each alarm being examined by a scientist. At the same time, we would like to avoid as many ``false alarms" as possible, since each alarm requires scientist effort. 

To formalize the example, our setting assumes two training sets: a clean training dataset involving a finite set of categories and a contaminated dataset that contains a fraction $\alpha$ of aliens. Our first contribution is to show that, in this setting, theoretical guarantees are possible given knowledge of an upper bound on $\alpha$. In particular, we give an algorithm that uses this knowledge to provide Probably Approximately Correct (PAC) guarantees for achieving a user-specified alien detection rate. While knowledge of a non-trivial upper bound on $\alpha$ may not always be possible, in many situations it will be possible to select a reasonable value based on domain knowledge, prior data, or by inspecting a sample of the test data.

The key idea behind our algorithm is to leverage modern anomaly detectors, which are trained on the clean data. Our algorithm combines the anomaly-score distributions over the clean and contaminated training data in order to derive an alarm threshold that achieves the desired guarantee on the alien detection rate on new test queries. In theory the detection rate guarantee will be met regardless of the quality of the anomaly detector. The quality of the detector, however, has a significant impact on the false alarm rate, with better detectors leading to fewer false alarms. 

We carry out experiments\footnote{Code for reproducing our experiments can be found at https://github.com/liusi2019/ocd.} on synthetic and benchmark datasets using a state-of-the-art anomaly detector, the Isolation Forest \cite{liu2008}. We vary the amount of training data, the fraction $\alpha$ of alien data points, along with the accuracy of the upper bound on $\alpha$ provided to our algorithm. The results indicate that our algorithm can achieve the guaranteed performance  when enough data is available, as predicted by the theory. The results also show that for the considered benchmarks, the Isolation Forest anomaly detector is able to support non-trivial false positive rates given enough data. The results also illustrate the inherent difficulty of the problem for small datasets and/or small values of $\alpha$. Overall, our results provide a useful baseline for driving future work on open category detection with guarantees. 


\section{Related Work}
Open category detection is related to the problem of one-class classification, which aims to detect outliers relative to a single training class. One-class SVMs (OCSVMs) \cite{Scholkopf:2001:ESH:1119748.1119749} are popular for this problem. However, they have been found to perform poorly for open category detection due to poor generalization \cite{Zhou2003}, which has been partly addressed by later work \cite{Manevitz:2002:OSD:944790.944808,4760149,1301575,6248047}. OCSVMs have been employed in a multi-class setting similar to open category detection \cite{6374555,Pritsos2013}. However, there are no direct mechanisms to control the alien detection rate of these methods, which is a key requirement for our problem setting. 

Work on classification with rejection/abstaining options \cite{1054406,Wegkamp2007LassoTC,TAX20081565,Pietraszek:2005:OAC:1102351.1102435,Geifman2017} allows classifiers to abstain from making predictions when they are not confident. While loosely related to open category detection, these approaches do not directly consider the possibility of novel categories, but rather focus on assessing confidence with respect to the known categories. Due to their closed-world discriminative nature,  it is easy to construct scenarios where such methods are incorrectly confident about the class of an alien and do not abstain. 

A variety of prior work has addressed variants of open category detection. This includes work on formalizing the concept of ``open space" to characterize the region of the feature space outside of the support of the training set \cite{6365193}. Variants of SVMs have also been developed, such as the One-vs-Set Machine \cite{6365193} and the Weibull-calibrated SVM \cite{6809169}. Additional work has addressed open category detection by tuning the decision boundary based on unlabeled data which contains data from novel categories \cite{Da:2014:LAC:2892753.2892797}. Approaches based on nearest neighbor methods have also been proposed \cite{MendesJÃºnior2017}. None of these methods, however, allow for the direct control of alien detection rates, nor do they provide theoretical guarantees. 
    
There is also recent interest in open category detection for deep neural networks applied to vision and text classification \cite{7780542,DBLP:journals/corr/abs-1709-08716}. These methods usually train a neural network in a standard closed-world setting, but then analyze various activations in the network in order to detect aliens. Another related line of work is detection of out-of-distribution instances,  which is similar to open category detection but assumes that the test data come from a completely different distribution compared to the training distribution \cite{hendrycks17baseline,liang2018enhancing}. All of this work is quite specialized to deep neural networks and does not provide direct control of alien detection rates or theoretical guarantees. 

\section{Problem Setup}
We consider open category detection where there is an unknown nominal data distribution $D_0$ over labeled examples from a known set of category labels. We receive as input a ``clean" nominal training set $S_0$ containing $k$ i.i.d.\ draws from $D_0$. In practice, $S_0$ will correspond to some curated labeled data that contains only known categories of interest. 

We also receive as input an unlabeled ``mixture" dataset $S_m$ that contains $n$ points drawn i.i.d.\ from a mixture distribution $D_m$. Specifically, the mixture distribution $D_m$ is a combination of the nominal distribution $D_0$ and an unknown alien distribution $D_a$, which is a distribution over novel categories (alien data points). We assume that $D_a$ is stationary, so that all alien points that appear as future test queries will also be drawn from $D_a$. 

At training time, we assume that $D_m$ is a mixture distribution, with probability $\alpha$ of generating an alien data point from $D_a$ and probability of $1-\alpha$ of generating a nominal point. Our results hold even if the test queries come from a mixture with a different value of $\alpha$ as long as the alien test points are drawn from $D_a$.

Given these datasets, our problem is to label test instances from $D_m$ as either ``alien" or ``nominal". In particular, we wish to achieve a specified \emph{alien detection rate}, which is the fraction of alien data points in $D_m$ that are classified as ``alien" (e.g., 95\%). At the same time we would like the \emph{false positive rate} to be small, which is the fraction of nominal data points incorrectly classified as aliens. 

Our approach to this problem assumes the availability of an anomaly detector that is trained on $S_0$ and assigns anomaly scores to all data points in both $S_0$ and $S_m$. Intuitively, the anomaly scores order the test examples according to how anomalous they appear relative to the nominal data (higher scores being more anomalous). An ideal detector would rank all alien data points higher than all nominals, though in practice, the ordering will not be so clean. Our approach labels data in $S_m$ by selecting a threshold on the anomaly scores and labeling all data points with scores above the threshold as aliens and the remaining points as nominals. Our key challenge is to select a threshold that provides a guarantee on the alien detection rate. 

\section{Algorithms for Open Category Detection}
In order to obtain theoretical guarantees, our algorithm assumes knowledge of the alien mixture probability $\alpha$ that generates the mixture data $S_m$. Later, we will show that knowing an upper bound on $\alpha$ is sufficient to obtain a guarantee. 

Our approach is based on considering the cumulative distribution functions (CDFs) over anomaly scores of a fixed anomaly detector. Let $F_0, F_a$, and $F_m$ be the CDFs of anomaly scores for the nominal data distribution $D_0$, alien distribution $D_a$, and mixture distribution $D_m$ respectively. Since $D_m$ is a simple mixture of $D_0$ and $D_a$, we can write $F_m$ as
\[F_{m}(x) = (1-\alpha)F_{0}(x) + \alpha F_{a}(x).\]
From this we can derive the CDF for $F_{a}$ in terms of $F_m$ and $F_0$:
\[F_{a}(x) = \frac{F_{m}(x) - (1-\alpha)F_{0}(x)}{\alpha}.\]
Given the ability to derive $F_a$, it is straightforward to achieve an alien detection rate of $1-q$ (e.g. 95\%) by selecting an anomaly score threshold $\tau_q$ that is the $q$ quantile of $F_a$ and raising an alarm on all test queries whose anomaly score is greater than $\tau_q$.

In reality, we do not have access to $F_m$ or $F_0$ and hence cannot exactly determine $F_a$. Rather, we have samples $S_m$ and $S_0$. Thus, our algorithm works with the empirical CDFs $\hat{F}_{0}$ and $\hat{F}_{m}$, which are simple step-wise constant approximations, and estimates an empirical CDF over aliens: 
\begin{equation}\label{Fa_est}
\hat{F}_{a}(x) = \frac{\hat{F}_{m}(x) - (1-\alpha)\hat{F}_{0}(x)}{\alpha}.
\end{equation}
Our algorithm computes the above estimate of $\hat{F}_{a}$ and uses it to select a threshold $\hat{\tau}_q$ to be the largest threshold such that $\hat{F}_{a}(\hat{\tau}_q) \leq q$, where $1-q$ is the target alien detection rate. This choice will minimize the number of false alarms. 
\label{headings}
The steps of this algorithm are as follows.\vspace{1em}

\begin{algorithm}
\caption{}
\begin{algorithmic}[1]
\STATE Get anomaly scores for all points in $S_{0}$ and $S_{m}$, denoted $x_1, x_2, \ldots, x_k$ and $y_1, y_2, \ldots, y_n$ respectively.
\STATE Compute empirical CDFs $\hat{F}_{0}$ and $\hat{F}_{m}$.
\STATE Calculate $\hat{F}_{a}$ using equation 1.
\STATE Output detection threshold $$\hat{\tau}_q = \max\{u \in S: \hat{F}_{a}(u)\leq q\},$$ where $S = \{x_1, x_2, \ldots, x_k, y_1, y_2, \ldots, y_n\}$. 
\end{algorithmic}
\end{algorithm}
Although $\hat{F}_m$ and $\hat{F}_0$ are both legal CDFs, the estimate for $\hat{F}_{a}$ from step 3 may not be a legal CDF, because it is the difference of two noisy estimates---it may not increase monotonically and it may even be negative. A good technique for dealing with this problem is to employ isotonization \cite{Barlow1972} and clipping. Isotonization finds the monotonically increasing function $\hat{F}_a^{*}$ closest to $\hat{F}_a$ in squared error. To convert $\hat{F}_{a}$ into a legal CDF, define $\check{F}_a = \min \{\max \{\hat{F}_a^{*},{\bf 0}\},{\bf 1}\}$, where the min and max operators are applied pointwise to their arguments. We performed experiments (shown in the supplementary materials) to test whether using $\check{F}_a$ in Step 4 would improve the performance of the overall algorithm. We found that it did not.

\section{Finite Sample Guarantee}

In the limit of infinite data (both nominal and mixture) and perfect knowledge of $\alpha$, $\hat{F}_a$ will converge to the true alien CDF, and our algorithm will achieve the desired alien detection rate. In this section, we consider the finite data case where $|S_0| = |S_m| = n$.  We derive a value for the sample size $n$ that guarantees with high probability over random draws of $S_0$ and $S_m$, that fraction $1-q-\epsilon$ of the alien test points will be detected, where $\epsilon$ is an additional error incurred because of the finite sample size $n$. 

Our key theoretical tool is a finite sample result on the uniform convergence of empirical CDF functions \cite{MASSART}. To use this result, we make the reasonable technical assumption that the nominal and alien CDFs, $F_0$ and $F_a$, are continuous.
In the following, let $\eta$ be the target alien detection rate, $q$ be the input to Algorithm 1, $\hat{\tau}_q$ be the estimated $q$-quantile of the alien CDF (step 4 of Alg. 1), and $\epsilon$ be an error parameter. The following theorem gives the sample complexity for guaranteeing that $1-\eta$ of the alien examples will be detected using threshold $\hat{\tau}_q$. 

\begin{theorem} 
Let $S_0$ and $S_m$ be nominal and mixture datasets containing $n$ i.i.d.\ samples from the nominal and mixture data distributions respectively. For any $\epsilon \in (0, 1-q)$ and $\delta \in (0,1)$, if $$n > \frac{1}{2}\ln{\frac{2}{1-\sqrt{1-\delta}}}\left(\frac{1}{\epsilon}\right)^2\left(\frac{2-\alpha}{\alpha}\right)^2,$$ then with probability at least $1-\delta$, Algorithm 1 will return a threshold $\hat{\tau}_q$ that achieves an alien detection rate of at least $1-\eta$, where $\eta = q + \epsilon$.
\end{theorem}
The proof is in the Appendix. Note that $n$ grows as $O(\frac{1}{\epsilon^2\alpha^2}\log{\frac{1}{\delta}})$. Hence, this guarantee is polynomial in all relevant parameters, which we believe is the first such guarantee for open category detection.
The result can be generalized to the case where $n_0 < n_m$; in practice, the larger the mixture sample $S_m$ is, the easier it is to estimate $\tau_q$, because this provides more alien points for estimating the $q$-th quantile of $F_a$.

The theorem gives us flexibility in setting $\epsilon$ and $q$ (the algorithm input) to achieve a guarantee of $1-\eta$. The $\epsilon$ parameter controls a trade-off between sample size and false alarm rate. To minimize the false alarm rate, we want to make $q$ large (to obtain a larger threshold), so we want to set $q$ close to $\eta$. But, as $q \rightarrow \eta$, $\epsilon \rightarrow 0$, and $n \rightarrow \infty$. To minimize the sample size $n$, we want to make $q$ as small as possible, because that allows $\epsilon$ to be larger and hence $n$ becomes smaller. The optimal setting of $\epsilon$ depends on how the false alarm rate grows with $\tau_q$, which in turn depends on the relative shape of $F_0$ and $F_a$. In a real safety application, we can estimate these from $S_0$ and $S_m$ and choose an appropriate $q$ value.

What if we don't know the exact value of $\alpha$? If our algorithm uses an upper bound $\alpha'$ on the true $\alpha$ to compute $\hat{F}_a$, we can still provide a guarantee. In this case, in addition to the assumptions in Theorem 1, we need a concept of an anomaly detector being \emph{admissible}. We say that an anomaly detector is \emph{admissible} for a problem, if the anomaly score CDFs satisfy $F_0(x)\geq F_m(x)$ for all $x \in \R$. Most reasonable anomaly detectors will be admissible in this sense, since the alien CDF will typically concentrate more mass toward larger anomaly score values compared to $F_0$. Indeed, if this is not the case, there is little hope since there is effectively no signal to distinguish between aliens and nominals. 
\begin{corollary}
Consider running Algorithm 1 using an upper bound $\alpha'$ on the true $\alpha$.  Under the same assumptions as Theorem 1, if the anomaly detector is admissible and$$n > \frac{1}{2}\ln{\frac{2}{1-\sqrt{1-\delta}}}\left(\frac{1}{\epsilon}\right)^2\left(\frac{2-\alpha'}{\alpha'}\right)^2,$$ then with probability at least $1-\delta$, Algorithm 1 will return a threshold $\hat{\tau}_q$ that achieves an alien detection rate of at least $1-\eta$, where $\eta=q + \epsilon$.
\end{corollary}
\noindent
The proof is in the Appendix. While we can achieve a guarantee using an upper bound on $\alpha'$, the returned threshold will be more conservative (smaller) than if we had used the true $\alpha$. This will result in higher false alarm rates, since more nominal points will be above the threshold. Thus it is desirable to use a value of $\alpha'$ that is as close to $\alpha$ as possible. 


\section{Experiments}
We performed experiments to answer four questions. Question Q1: how accurate is our estimate of $\hat{\tau}_q$ as a function of $n$ and $\alpha$? Question Q2: how loose are the bounds from Theorem 1? Question Q3: what are typical values of the false alarm rates for various settings of $n$ and $\alpha$ on real datasets? Question Q4: how do these observed values change if we employ an overestimate $\alpha'>\alpha$?

All of our experiments employ the Isolation Forest anomaly detector \cite{liu2008}, which has been demonstrated to be a state-of-the-art detector in recent empirical studies \cite{emmott2013}. In the Supplementary Materials we show similar results with the LODA anomaly detector \cite{Pevny2015}. 

To address Q1 and Q2, we run controlled experiments on synthetic data. The data points are generated from 9-dimensional normal distributions. The dimensions of the nominal distribution $D_0$ are independently distributed as $N(0,1)$. The alien distribution is similar, but with probability 0.4, 3 of the 9 dimensions (chosen uniformly at random) are distributed as $N(3,1)$ and with probability 0.6, 4 of the 9 dimensions (chosen uniformly at random) follow $N(3,1)$. This ensures that the anomalies are not highly similar to each other and models the situation in which there are many different kinds of alien objects, not just a single alien class forming a tight cluster.

\begin{figure}[h]
\centering
\includegraphics[width=\columnwidth]{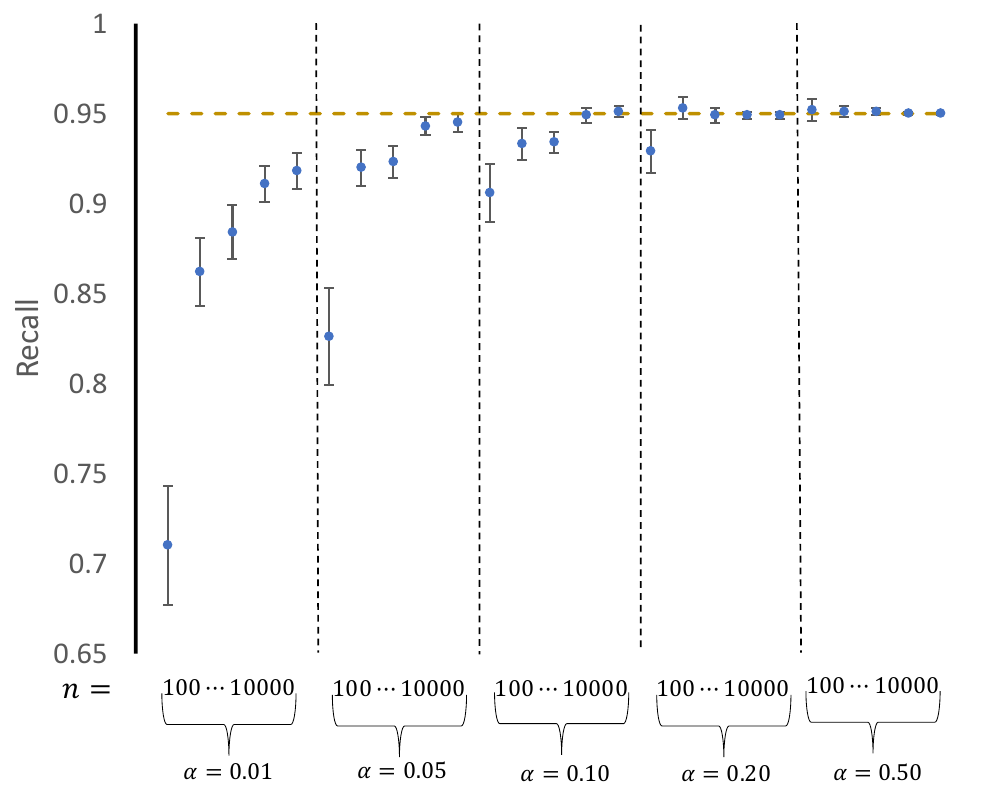}
\vspace*{-0.2in}
\caption{Comparison of recall achieved by $\hat{\tau}_q$ compared to oracle recall of 0.95. Error bars are 95\% confidence intervals. Settings of $n$ and $\alpha$ increase from left to right starting with $\alpha=0.01$ and $n \in \{100, 500, 1\text{K}, 5\text{K}, 10\text{K}\}$ up to $\alpha=0.5$ and $n=10\text{K}$.}
\label{fig:tau}
\end{figure}
In each experiment, the nominal dataset and the mixture dataset are of the same size $n$, and the mixture dataset contains a proportion $\alpha$ of anomaly points. We fixed the target quantile to be $q=0.05$. 
The experiments are carried out for $n \in \{100, 500, 1\text{K}, 5\text{K}, 10\text{K}\}$ and $\alpha \in \{ 0.01,0.05,0.10,0.20,0.50\}$. 
For testing, we create two large datasets $G_0$ and $G_a$, with $G_0$ being a pure nominal dataset, $G_a$ being a pure alien dataset, and $|G_0| = |G_a| = 20K$. The Isolation Forest algorithm computes $1000$ full depth isolation trees on the nominal data. Each tree is grown on a randomly-selected 20\% subsample of the clean data points. We compute anomaly scores for the nominal points via out-of-bag estimates and anomaly scores for the mixture points, $G_0$, and $G_a$ using the full isolation forest.  For each combination of $n$ and $\alpha$, we repeat the experiment $100$ times. We measure the fraction of aliens detected (the ``recall'') and the fraction of nominal points declared to be alien (the ``false positive rate'') by applying the $\hat{\tau}_q$ estimate to threshold the anomaly scores in $G_0$ and $G_a$.  

\begin{figure}[t]
\centering
\includegraphics[width=\columnwidth]{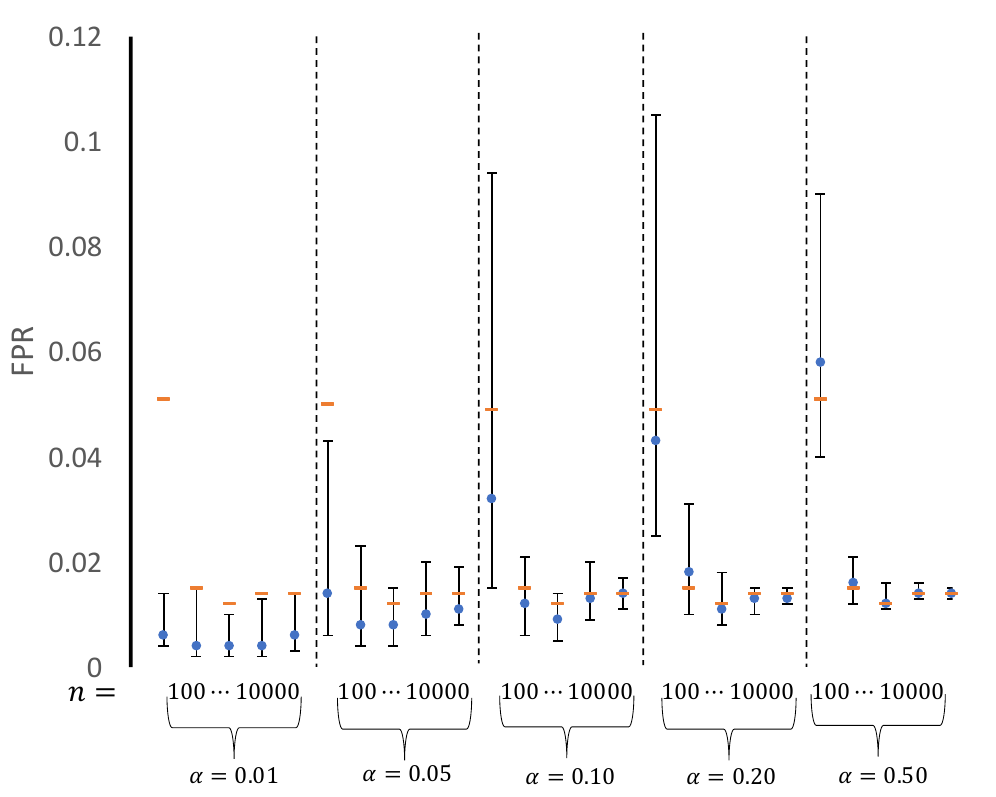}
\vspace*{-0.2in}
\caption{Comparison of oracle FPR to the FPR achieved by $\hat{\tau}_q$. Error bars span from the 25th to 75th percentile with the blue dot marking the median of the 100 trials. Orange markers indicate the oracle FPR.  Settings of $n$ and $\alpha$ increase from left to right starting with $\alpha=0.01$ and $n \in \{100, 500, 1\text{K}, 5\text{K}, 10\text{K}\}$ up to $\alpha=0.5$ and $n=10\text{K}$.}
\label{fig:fpr}
\end{figure}

To assess the accuracy of our $\hat{\tau}_q$ estimates (Q1), we could compare them to the true values. However, this comparison is hard to interpret, because $\tau$ is expressed on the scale of anomaly scores, which are somewhat arbitrary. Instead, Figure~\ref{fig:tau} plots the recall achieved by $\hat{\tau}_q$. If $\hat{\tau}_q$ had been estimated perfectly, the recall would always be $1 - q = 0.95$. However, we see that the recall is often less than 0.95, which indicates that $\hat{\tau}_q$ is over-estimated, especially when $n$ and $\alpha$ are small. This behavior is predicted by our theory, where we see that the sample size requirements grow inversely with $\alpha^2$. For larger $\alpha$ and $n$, the recall guarantee is generally achieved. Figure~\ref{fig:fpr} compares the false positive rate of the true oracle $\tau_q$ to the false positive rate of the estimate $\hat{\tau}_q$. For each combination of $\alpha$ and $n$, we have 100 replications of the experiment and therefore 100 estimates $\hat{\tau}_a$ and 100 FPR rates. For each of these, the true FPR is computed using $G_0$. The  error bars summarize the resulting 100 FPR values by the median and inter-quartile range. We see that for small $n$ and $\alpha$, the FPR can be quite different from the oracle rate, but for larger $n$ and $\alpha$, the estimates are very good.  

\begin{figure}[b!]
\centering
\includegraphics[width=\columnwidth]{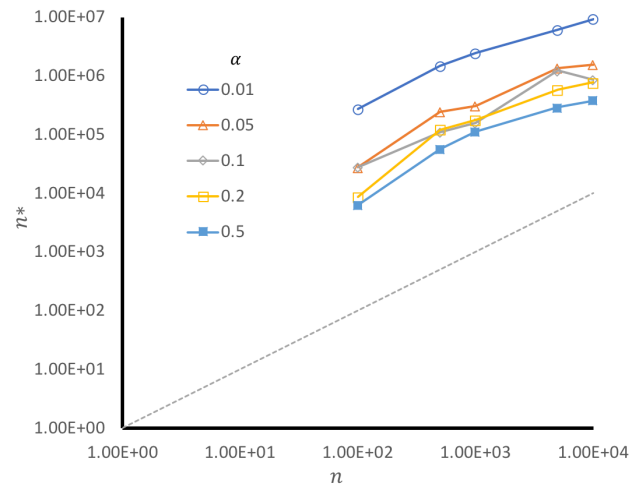}
\vspace*{-0.2in}
\caption{The log sample size $n^*$ required by Theorem 1 in order to guarantee the actual observed recall versus the log actual sample size $n$.}
\label{fig:n}
\end{figure}

To assess the looseness of the bounds (Q2), for each combination of $n$ and $\alpha$, we fix $\delta=0.05$ and compute the value of $\eta$ such that 95 of the 100 runs achieved a recall of at least $1-\eta$ (thus $\eta$ empirially achieves the $1-\delta$ guarantee). We then compute $\epsilon = \eta - q$ and the corresponding required sample size $n^*$ according to Theorem 1. Figure~\ref{fig:n} shows a plot of $n^*$ versus the actual $n$. The distance of these points from the $n^*= n$ diagonal line show that the theory is fairly loose, although it becomes tighter as $n$ gets large.

\begin{figure}[ht]
\centering
\includegraphics[width=\columnwidth]{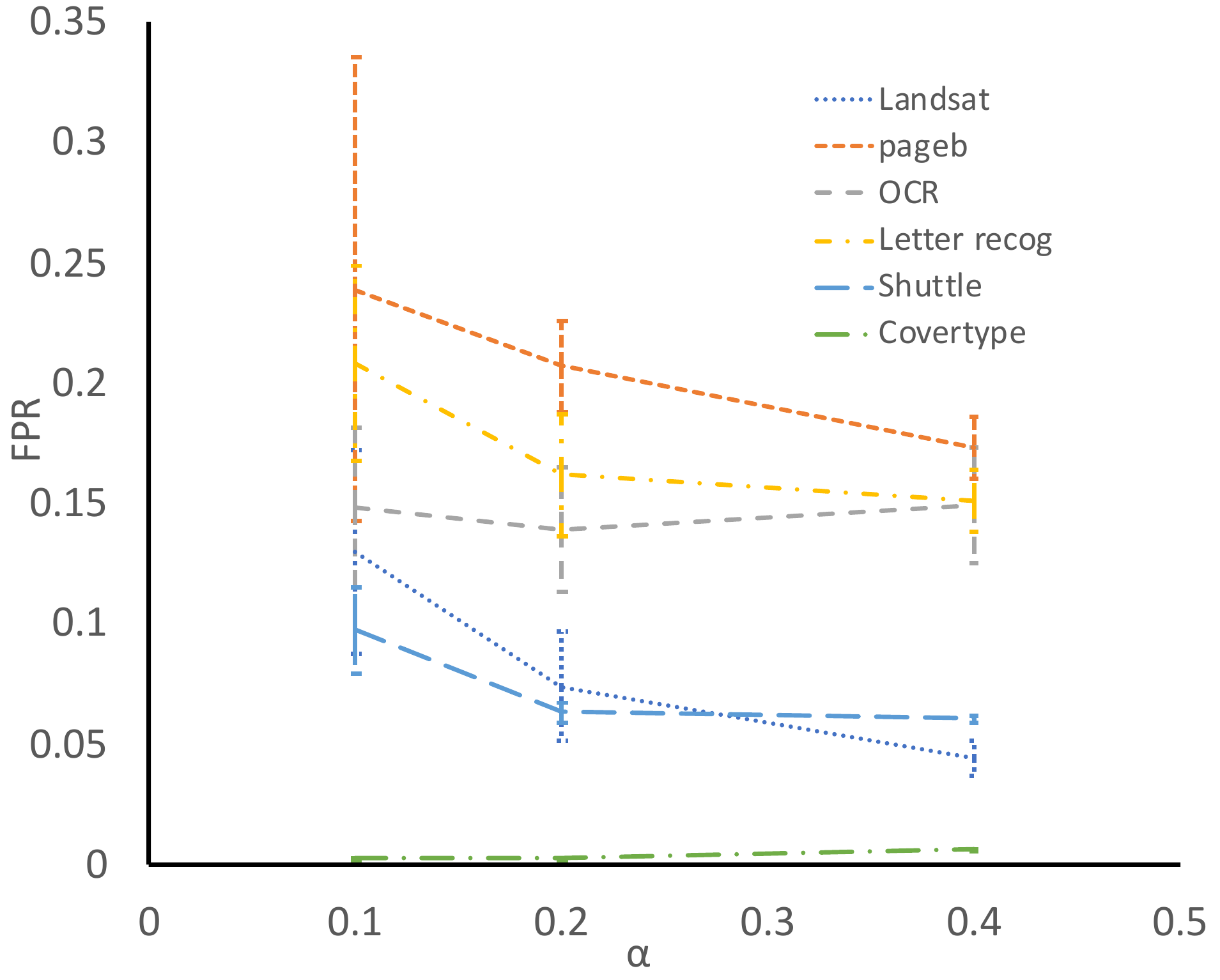}
\vspace*{-0.2in}
\caption{False positive rates on six UCI datasets as a function of $\alpha$ ($q=0.05$, $\delta = 0.05$).}
\label{fig:FAR}
\end{figure}

\begin{figure}[b!]
\centering
\includegraphics[width=\columnwidth]{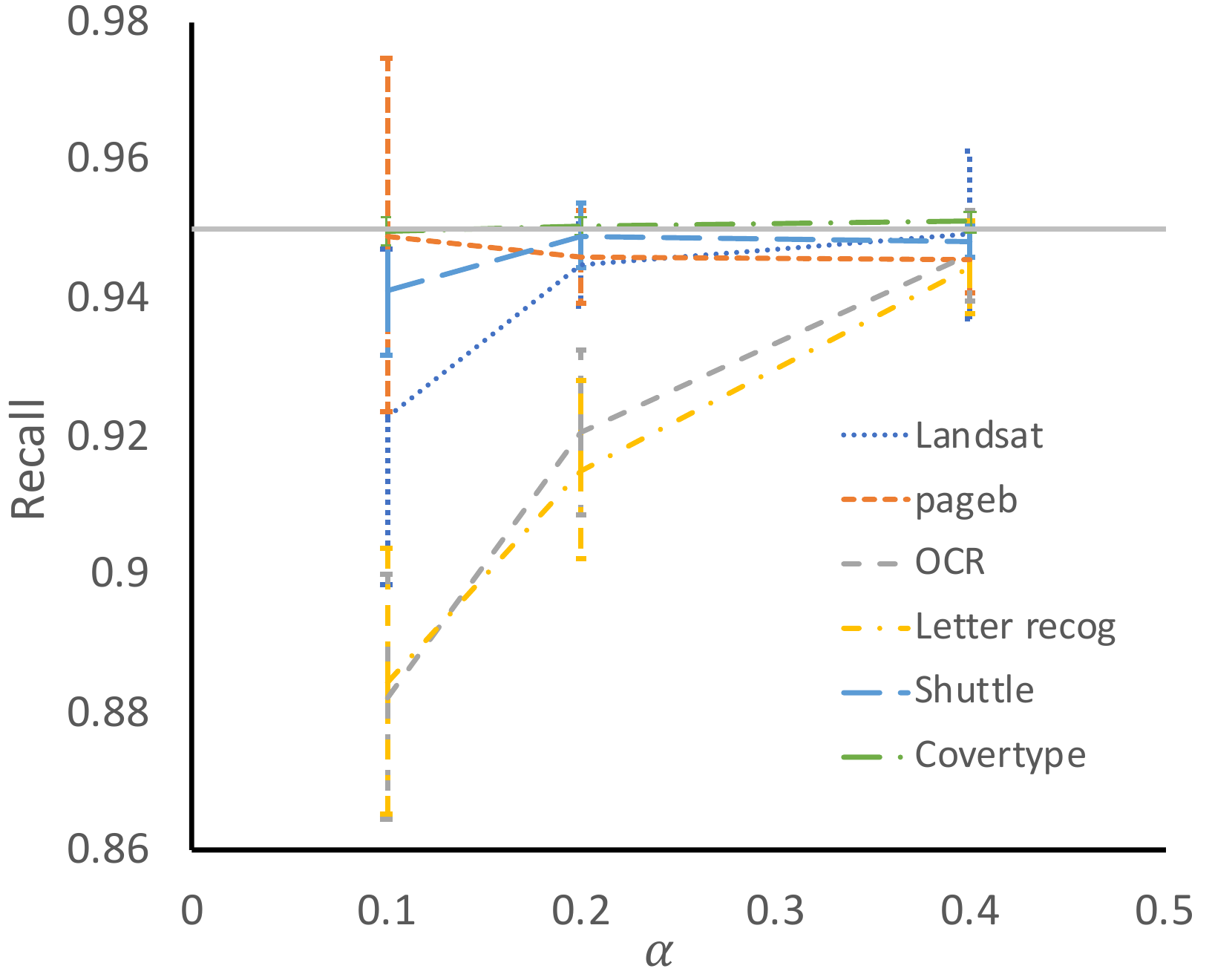}
\vspace*{-0.2in}
\caption{Recall rates on six UCI datasets as a function of $\alpha$ ($q=0.05$, $\delta = 0.05$)}
\label{fig:Recall}
\end{figure}

\begin{figure}[ht]
\centering
\includegraphics[width=\columnwidth]{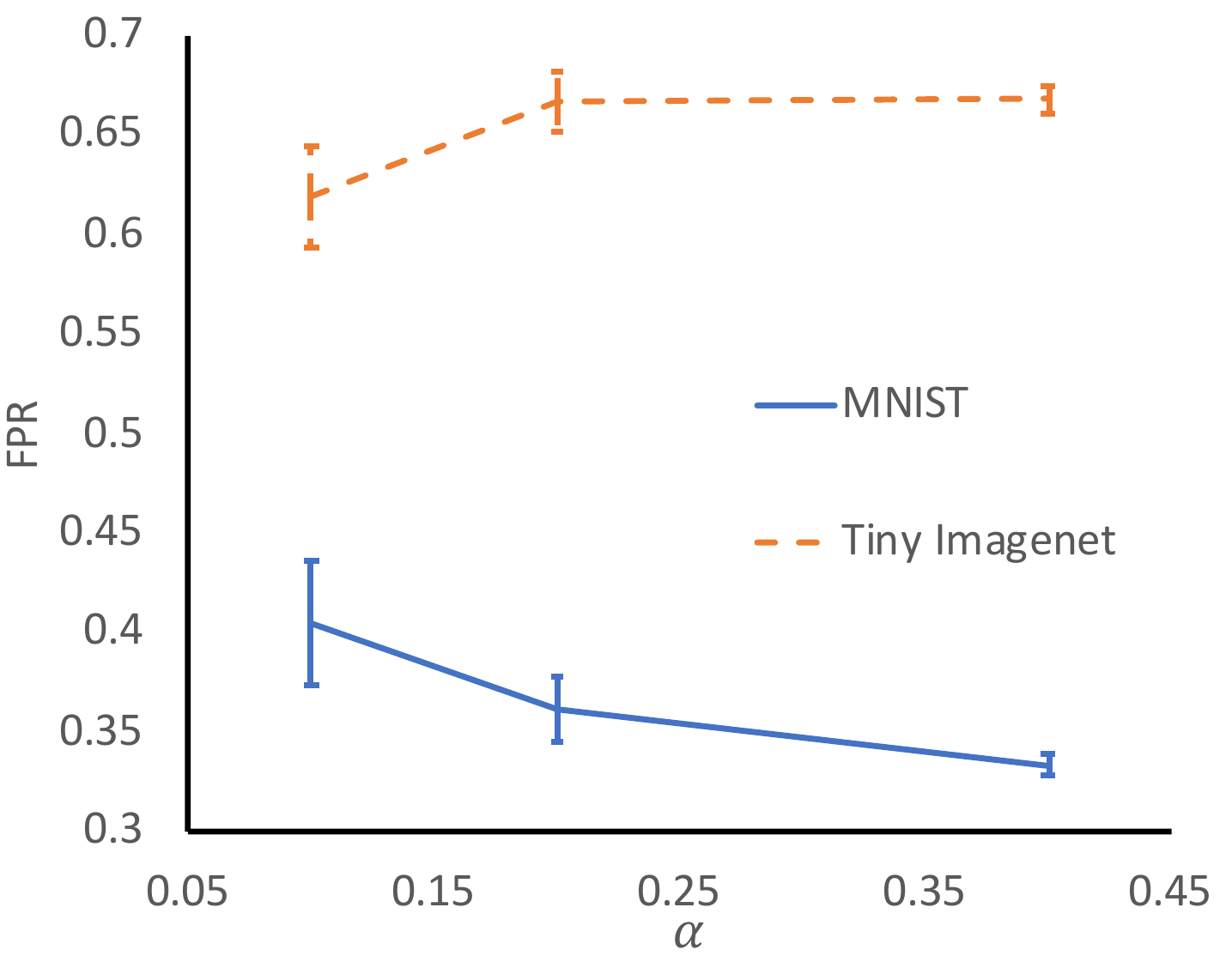}
\vspace*{-0.2in}
\caption{False positive rates on two image datasets as a function of $\alpha$ ($q = 0.05, \delta = 0.05$).}
\label{fig:image-fpr}
\end{figure}

\begin{figure}[ht]
\centering
\includegraphics[width=\columnwidth]{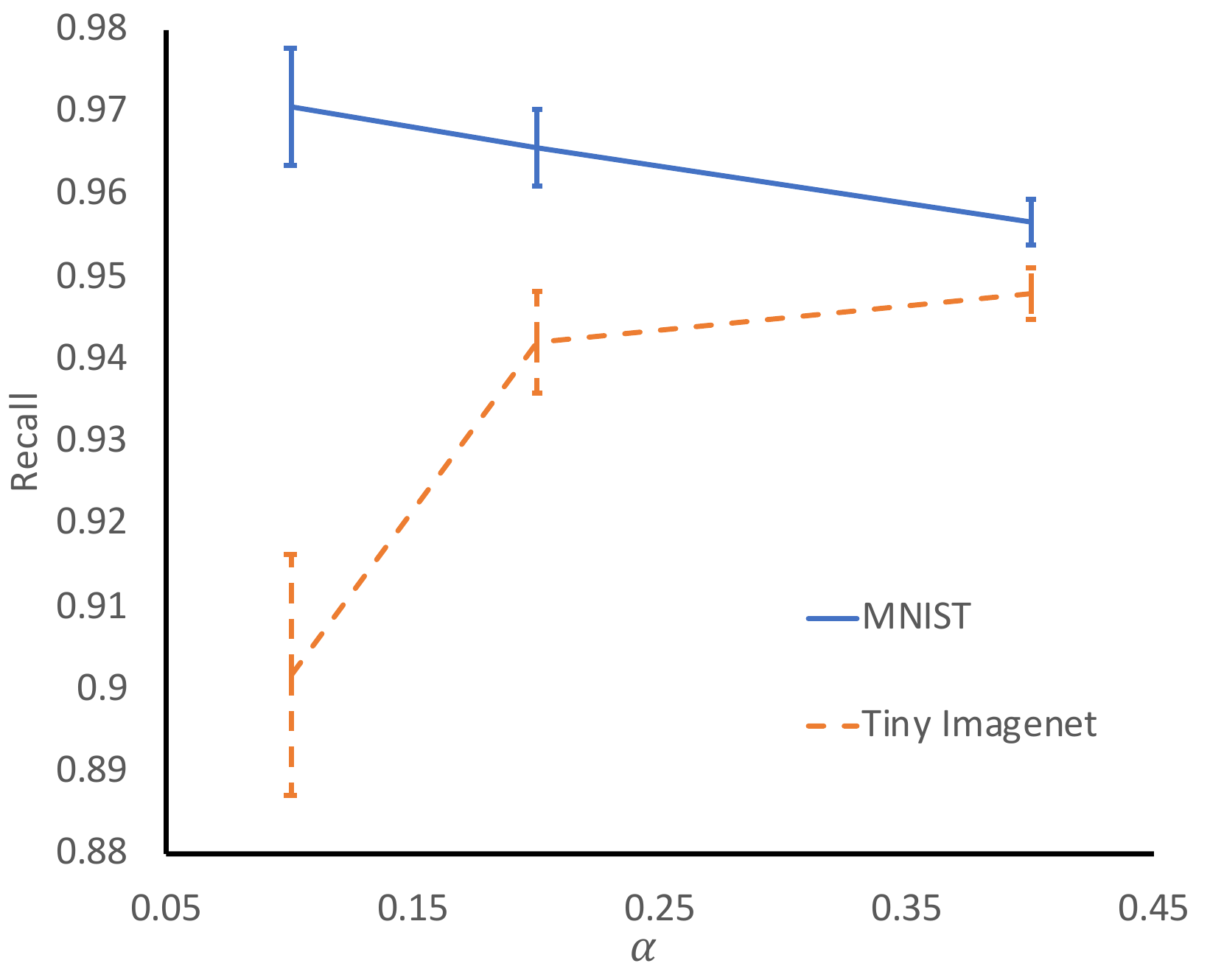}
\vspace*{-0.2in}
\caption{Recall rates on two image datasets as a function of $\alpha$ ($q = 0.05, \delta = 0.05$).}
\label{fig:image-recall}
\end{figure}

\begin{figure}[ht]
\centering
\includegraphics[width=\columnwidth]{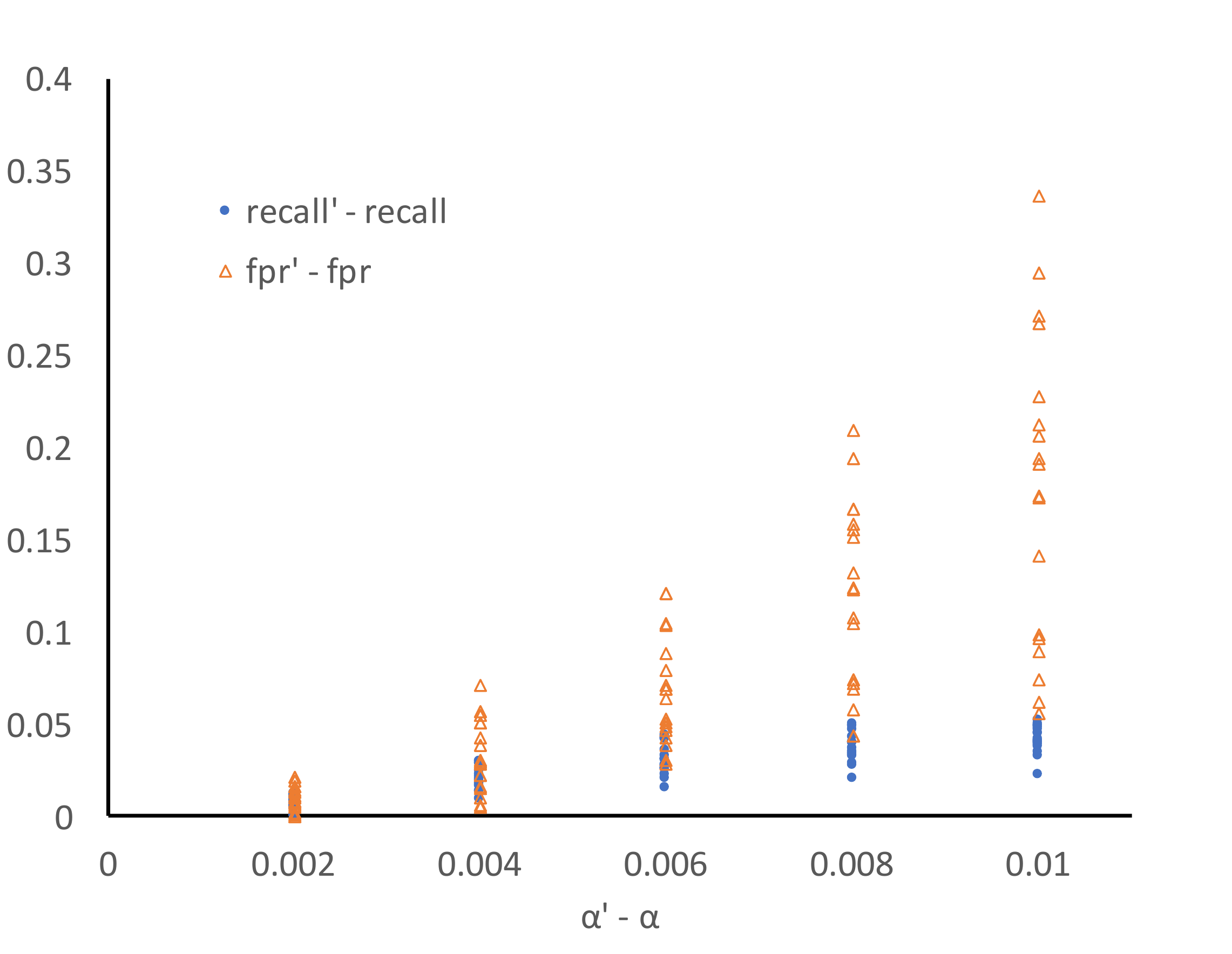}
\vspace*{-0.2in}
\caption{Change in recall and false positive rate as a function of $\alpha'-\alpha$ for six UCI datasets; $\alpha \in \{0.1, 0.2, 0.4\}$}
\label{fig:recall-fpr-alpha}
\end{figure}

\textbf{Benchmark Data Experiments.} To address our third and fourth questions, we performed experiments on six UCI multiclass datasets: Landsat, Opt.digits, pageb, Shuttle, Covertype and MNIST. In addition to these, we provide results for the Tiny ImageNet dataset. In each multiclass dataset, we split the classes into two groups: nominal and alien. For Tiny ImageNet, we train a deep neural network classifier on 200 nominal classes and treat the remaining 800 as aliens. The nominal classes for UCI datasets are MNIST(1,3,7), Landsat(1,7), OCR(1,3,4,5,7), pageb(1,5), Letter  recognition(1,3), and Shuttle(1,4). We generated nominal and mixture datasets for various values of $\alpha$. The value of $n$ for each dataset is 1532 for Landsat,788 for Letter recognition, 568 for OCR, 4912 for pageb, 5000 for Shuttle, 13,624 for Covertype, 11,154 for MNIST, and 10,000 for Tiny ImageNet. Because we cannot create datasets with large $n$, we cannot measure the true value of $\tau_q$. 

After computing the anomaly scores for both nominal and mixture datasets, we applied Algorithm 1 within a 10-fold cross validation. We divide the mixture data points at random into 10 groups. For each fold, we estimate $\hat{F}_a$ and $\hat{\tau}_a$ from 9 of the 10 groups and then score the mixture points in the held-out fold according to $\hat{\tau}_a$. In all other respects, the experimental protocol is the same as for the synthetic data. For Tiny ImageNet, the anomaly scores are obtained by applying a baseline method \cite{hendrycks17baseline}.

To answer Q3, Figures~\ref{fig:FAR} and \ref{fig:image-fpr} plot the false positive rate as a function of $\alpha$ for the UCI and vision datasets, respectively. We see that the FPR ranges from 3.6\% to 26.9\% on UCI depending on the dataset and the level of $\alpha$. The vision datasets have higher FPR, especially MNIST, which has a large number of alien classes that are not distinguished well by the anomaly detector. The FPR depends primarily on the domain, because the key issue is how well the anomaly detector distinguishes between nominal and alien examples. The false alarm rate generally improves as $\alpha$ increases. In some applications, it may be possible to enrich $S_m$ so that $\alpha$ is larger on the training set to take advantage of this phenomenon. It is interesting to note that once $\hat{\tau}_a$ has been computed, it can be applied to test datasets having different (or unknown) values of $\alpha$.

Figures~\ref{fig:Recall} and \ref{fig:image-recall} plot the recall rate as a function of $\alpha$ for the UCI and vision datasets. We set $q=0.05$ in these experiments. Theorem 1 only guarantees a recall of $1- q-\epsilon$, where $\epsilon$ depends on $n$. Hence, it is nice to see that for three of the domains (Shuttle, Covertype, and Landsat) in UCI and for both vision datasets, the recall is very close to $1-q = 0.95$.
These are the domains with the largest values of $n$. The value of $\alpha$ has a bigger impact on recall than it does on FPR. This is because the effective number of alien training examples is $\alpha n$, which can be very small for some datasets when $\alpha=0.1$. This shows that in applications such as fraud detection, where $\alpha$ may be very small, the mixture dataset $S_m$ needs to be very large.

To answer Q4 regarding the impact of using an incorrect value $\alpha'>\alpha$, we repeated these experiments with $\alpha' = \alpha + \xi$, for $\xi \in \{0.002, 0.004, 0.006, 0.008, 0.010\}$. Figure~\ref{fig:recall-fpr-alpha} plots the change in false positive rate and recall as a function of $\alpha' - \alpha$. Two points are plotted for each combination of $\alpha'$ and dataset, the change in Recall and the change in FPR. We observe that the recall increases slightly (in the range from 0.01 to 0.05). However, the false positive rate increases by much larger amounts (from 0.01 to 0.336). This demonstrates that it is very important to determine the value of $\alpha$ accurately. 

\section{Summary}

We have taken a step toward open category detection with guarantees by providing a PAC-style guarantee on the probability of detecting $1-\eta$ of the aliens on the test data. This is the first such guarantee under any similarly general conditions. We have shown that this guarantee is satisfied in our experiments, although the guarantee is somewhat loose, especially on small training sets. Obtaining a guarantee requires more data than standard PAC guarantees on expected prediction accuracy. This is because we must estimate the $q$ quantile of the alien anomaly score distribution, where $q$ is typically quite small. Nonetheless, our experiments show that our algorithm gives good recall performance and non-trivial false alarm rates on datasets of reasonable size. 

It is important to note that the very formulation of a PAC-style guarantee on the probability of detecting aliens requires assuming that the aliens are drawn from a well-defined distribution $D_a$. While this is appropriate in some applications, such as the insect survey application described in the introduction, it is not appropriate for adversarial settings. In such settings, a PAC-style guarantee does not make sense, and some other form of safety guarantee needs to be formulated.

To obtain the guarantee, we employ two training datasets: a clean dataset that contains no aliens and an (unlabeled) contaminated dataset that contains a known fraction $\alpha$ of aliens. An important theoretical problem for future research is to develop a method that can estimate a tight upper bound on $\hat{\alpha} > \alpha$. We believe this is possible, but we have not yet found a method that guarantees that $\hat{\alpha} > \alpha$.

Our guarantee requires more data as $\alpha$ becomes small. Fortunately, when $\alpha$ is small, it may be possible in some applications to afford lower recall rates, since the frequency of aliens will be smaller. However, in safety-critical applications where a single undetected alien poses a serious threat, there is little recourse other than to collect more data or allow for higher false positive rates. 

\section*{Acknowledgements} This research was supported by a gift from Huawei, Inc., and grants from the Future of Life Institute and the NSF Grant 1514550. Any opinions, findings, and conclusions or recommendations expressed in this material are those of the author(s) and do not necessarily reflect the views of the sponsors.
\appendix

\section{Proof for Theorem 1}
Suppose there are $n$ random variables which are i.i.d.\ from the distribution with CDF $F$ and let $\hat{F}_n$ be the empirical CDF calculated from this sample. Then Massart (\citeyear{MASSART}) shows that 
\begin{equation}\label{eq:1}
 P(\sqrt{n}  \sup_x |\hat{F}_n (x) - F(x)| > \lambda ) \leq 2 \exp (-2\lambda^2)
\end{equation}
holds without any restriction on $\lambda$. 
Making use of this, and assuming we use the same sample size $n$ for both the mixture dataset and the clean data	set, for any $\epsilon \in (0, 1-q)$, we seek to determine how large $n$ needs to be in order to guarantee that with probability at least $1-\delta$ our quantile estimate $\hat{\tau}_q$ satisfies $F_a(\hat{\tau}_q) \leq q + \epsilon$. To achieve this, we want to have 
\[P(\sup_x|\hat{F}_a(x) - F_a(x)|> \epsilon) \leq \delta.\]
We have 
\begin{eqnarray*}
& & P(\sup_x|\hat{F}_a(x) - F_a(x)|> \epsilon) \\
& = &P(\sup_x|\frac{\hat{F}_{m}(x) - (1-\alpha)\hat{F}_{0}(x)}{\alpha} - \\
& &\frac{F_{m}(x) - (1-\alpha)F_{0}(x)}{\alpha}|>\epsilon)\\
& = & P(\sup_x|\frac{1}{\alpha}(\hat{F}_{m}(x) - F_{m}(x)) - \\
& &\frac{1-\alpha}{\alpha}(\hat{F}_{0}(x) - F_{0}(x))|>\epsilon)\\
&\leq& P((\frac{1}{\alpha}\sup_x|\hat{F}_{m}(x) - F_{m}(x)| + \\
& & \frac{1-\alpha}{\alpha}\sup_x|\hat{F}_{0}(x) - F_{0}(x)| )> \epsilon)\\
&\leq& P(\{\frac{1}{\alpha}\sup_x|\hat{F}_{m}(x) - F_{m}(x)| > \frac{1}{2-\alpha}\epsilon\}\\
& &\cup\; \{\frac{1-\alpha}{\alpha}\sup_x|\hat{F}_{0}(x) - F_{0}(x)| > \frac{1-\alpha}{2-\alpha}\epsilon\})\\
& = & P(\{\sup_x|\hat{F}_{m}(x) - F_{m}(x)| > \frac{\alpha}{2-\alpha}\epsilon\}\\
& & \cup\; \{\sup_x|\hat{F}_{0}(x) - F_{0}(x)| > \frac{\alpha}{2-\alpha}\epsilon\}).
\end{eqnarray*}
Making use of (\ref{eq:1}), when
 \[n > \frac{1}{2}\ln{\frac{2}{1-\sqrt{1-\delta}}}(\frac{1}{\epsilon})^2(\frac{2-\alpha}{\alpha})^2,\]
we will have 
\begin{eqnarray*}
P(\sup_x|\hat{F}_{m}(x) - F_{m}(x)| > \frac{\alpha}{2-\alpha}\epsilon)\leq 1- \sqrt{1-\delta},\\
P(\sup_x|\hat{F}_{0}(x) - F_{0}(x)| > \frac{\alpha}{2-\alpha}\epsilon) \leq 1- \sqrt{1-\delta}.
\end{eqnarray*}
In this case we will have
\begin{eqnarray*}
& & P(\sup_x|\hat{F}_a(x) - F_a(x)|> \epsilon) \\
& \leq & 1- P(\{\sup_x|\hat{F}_{m}(x) - F_{m}(x)| \leq \frac{\alpha}{2-\alpha}\epsilon\}\\
& & \cap\; \{\sup_x|\hat{F}_{0}(x) - F_{0}(x)| \leq \frac{\alpha}{2-\alpha}\epsilon\})\\
&\leq &  1- (1 - 1 + \sqrt{1-\delta})^2\\
& = & \delta.
\end{eqnarray*}
Now we have with probability at least $1-\delta$, 
\[|\hat{F}_{a}(x) - F_{a}(x)| \leq \epsilon, \ \ \forall x \in \R.\]
If this inequality holds, then for any value $\hat{\tau}_q$ such that $\hat{F}_{a}(\hat{\tau}_q)\leq q $, we have 
\[F_{a}(\hat{\tau}_q) \leq \hat{F}_{a}(\hat{\tau}_q) + \epsilon \leq q+\epsilon.\]
So we have with probability at least $1-\delta$, any $\hat{\tau}_q$ satisfying $\hat{F}_{a}(\hat{\tau}_q)\leq q $ will satisfy $F_a(\hat{\tau}_q) \leq q+\epsilon$. \hfill{$\square$}

\section{Proof for Corollary 1}
If $\alpha' \geq \alpha$, and if we write 
\[ F'_{a}(x) = \frac{F_{m}(x) - (1-\alpha')F_{0}(x)}{\alpha'},\]
then $F'_{a}$ is still a legal CDF, because 
\[F'_{a}(-\infty) = 0, \ \ F'_{a}(\infty) = 1,\]
and it is easy to show that $F'_{a}$ is monotonically nondecreasing. 

But\[F'_{a}(x) - F_{a}(x) =  \frac{(\alpha - \alpha')(F_{m}(x) - F_{0}(x))}{\alpha\alpha'} \geq 0, \forall x \in \R,\]
and because of this, if we let $\hat{\tau}'_q$ denote the threshold we get from using $\alpha'$, we will have $F_a(\hat{\tau}'_q) \leq F'_a(\hat{\tau}'_q)$.
By the proof of previous theorem, we know that when $n > \frac{1}{2}\ln{\frac{2}{1-\sqrt{1-\delta}}}(\frac{1}{\epsilon})^2(\frac{2-\alpha'}{\alpha'})^2$, we have with probability at least $1-\delta$, $F'_a(\hat{\tau}'_q) \leq q + \epsilon$, and thus we have $F_a(\hat{\tau}'_q) \leq  q + \epsilon$.\hfill{$\square$}

\bibliography{example_paper}
\bibliographystyle{icml2018}

\newpage
\newpage
\clearpage
\appendix
\section{Experimental Results from Synthetic Datasets}
In this section we include the simulation results on synthetic datasets from using two different anomaly detectors, Isolation Forest and LODA in table 1-3 and 4-6 respectively. For using LODA, when training it on the nominal dataset, we build 1\,000 random projections, and each of them is built using a bootstrap resample of the nominal dataset. After finishing building all projections, we calculate the anomaly score for each point in nominal dataset only using the projections that didn't use this point, and calculate the anomaly scores for mixture dataset, $G_0$ and $G_a$ using all the projections. For all cases, we include results from targeting on different recalls which are $98\%$, $95\%$ and $90\%$. In table 1-6, the oracle FPR column is the mean of 100 oracle FPRs in each setting. 

In table 7, we include the results we used for plotting figure 2. The results are the 1st quartile, median and 3rd quartile of FPR from experiments using Iforest with target recall $95\%$. Here the oracle FPR column is the median of 100 oracle FPRs. 

\begin{table*}
\begin{center}
  \caption{$n^*$, recall (i.e. alien detection rate) and false positive rate from experiments using 9-dimensional normal data, $98\%$, iForest}
  \label{synthetic-table-3}
  \vskip 0.15in
  \centering
\begin{tabular}{lrrccc|ccc}
    \toprule
        & & \multicolumn{4}{c}{Basic CDF}&\multicolumn{3}{c}{Iso CDF}\\
      &   & \multicolumn{2}{c}{Recall} & \multicolumn{2}{c}{False Positive Rate}&\multicolumn{1}{c}{Recall} & \multicolumn{2}{c}{False Positive Rate} \\
    \cmidrule{2-9}
           $\alpha$   & $n$ & $n^*$ &Recall$\pm$CI&FPR$\pm$CI& Oracle& Recall$\pm$CI &FPR$\pm$CI & Oracle\\
    \midrule
0.01	&	100	&	247818	&	0.710$\pm$0.033	&	0.033$\pm$0.027	&	0.102	&	0.929$\pm$0.029	&	0.512$\pm$0.080	&	0.102	\\
	&	500	&	1167215	&	0.862$\pm$0.019	&	0.033$\pm$0.024	&	0.042	&	0.972$\pm$0.016	&	0.543$\pm$0.079	&	0.042	\\
	&	1000	&	1829649	&	0.884$\pm$0.015	&	0.031$\pm$0.024	&	0.036	&	0.980$\pm$0.009	&	0.574$\pm$0.080	&	0.036	\\
	&	5000	&	4236646	&	0.920$\pm$0.010	&	0.060$\pm$0.038	&	0.039	&	0.985$\pm$0.007	&	0.506$\pm$0.079	&	0.039	\\
	&	10000	&	6363404	&	0.932$\pm$0.009	&	0.065$\pm$0.034	&	0.037	&	0.984$\pm$0.007	&	0.520$\pm$0.080	&	0.037	\\
\bottomrule
  \end{tabular}
 \end{center}
\end{table*}

\begin{table*}
\begin{center}
  \caption{$n^*$, recall (i.e. alien detection rate) and false positive rate from experiments using 9-dimensional normal data, $95\%$, iForest}
  \label{synthetic-table-2}
  \vskip 0.15in
  \centering
  \begin{tabular}{lrrccc|ccc}
    \toprule
        & & \multicolumn{4}{c}{Basic CDF}&\multicolumn{3}{c}{Iso CDF}\\
      &   & \multicolumn{2}{c}{Recall} & \multicolumn{2}{c}{False Positive Rate}&\multicolumn{1}{c}{Recall} & \multicolumn{2}{c}{False Positive Rate} \\
    \cmidrule{2-9}
           $\alpha$   & $n$ & $n^*$ &Recall$\pm$CI&FPR$\pm$CI& Oracle& Recall$\pm$CI &FPR$\pm$CI & Oracle\\
    \midrule
0.01	&	100	&	275039	&	0.710$\pm$0.033	&	0.033$\pm$0.027	&	0.052	&	0.929$\pm$0.029	&	0.509$\pm$0.080	&	0.052	\\
	&	500	&	1474209	&	0.862$\pm$0.019	&	0.033$\pm$0.024	&	0.015	&	0.972$\pm$0.016	&	0.533$\pm$0.079	&	0.015	\\
	&	1000	&	2462157	&	0.884$\pm$0.015	&	0.030$\pm$0.024	&	0.012	&	0.978$\pm$0.010	&	0.557$\pm$0.081	&	0.012	\\
	&	5000	&	6171393	&	0.911$\pm$0.010	&	0.039$\pm$0.030	&	0.014	&	0.982$\pm$0.008	&	0.496$\pm$0.080	&	0.014	\\
	&	10000	&	9309633	&	0.918$\pm$0.010	&	0.054$\pm$0.032	&	0.014	&	0.981$\pm$0.008	&	0.495$\pm$0.081	&	0.014	\\
\bottomrule
  \end{tabular}
 \end{center}
\end{table*}

\begin{table*}
\begin{center}
  \caption{$n^*$, recall (i.e. alien detection rate) and false positive rate from experiments using 9-dimensional normal data, $90\%$, iForest}
  \label{synthetic-table-1}
  \vskip 0.15in
  \centering
  \begin{tabular}{lrrccc|ccc}
    \toprule
        & & \multicolumn{4}{c}{Basic CDF}&\multicolumn{3}{c}{Iso CDF}\\
      &   & \multicolumn{2}{c}{Recall} & \multicolumn{2}{c}{False Positive Rate}&\multicolumn{1}{c}{Recall} & \multicolumn{2}{c}{False Positive Rate} \\
    \cmidrule{2-9}
           $\alpha$   & $n$ & $n^*$ &Recall$\pm$CI&FPR$\pm$CI& Oracle& Recall$\pm$CI &FPR$\pm$CI & Oracle\\
    \midrule
0.01	&	100	&	331513	&	0.710$\pm$0.033	&	0.033$\pm$0.027	&	0.026	&	0.929$\pm$0.029	&	0.509$\pm$0.080	&	0.026\\
	&	500	&	2340744	&	0.862$\pm$0.019	&	0.033$\pm$0.024	&	0.005	&	0.970$\pm$0.016	&	0.517$\pm$0.078	&	0.005\\
	&	1000	&	3222506	&	0.859$\pm$0.014	&	0.011$\pm$0.008	&	0.004	&	0.976$\pm$0.011	&	0.542$\pm$0.081	&	0.004\\
	&	5000	&	5918805	&	0.869$\pm$0.011	&	0.012$\pm$0.017	&	0.004	&	0.976$\pm$0.01	&	0.476$\pm$0.079	&	0.004\\
	&	10000	&	12543171	&	0.884$\pm$0.010	&	0.012$\pm$0.009	&	0.005	&	0.971$\pm$0.011	&	0.458$\pm$0.080	&	0.005\\
\bottomrule
  \end{tabular}
 \end{center}
\end{table*}

\begin{table*}
\begin{center}
  \caption{$n^*$, recall (i.e. alien detection rate) and false positive rate from experiments using 9-dimensional normal data, $98\%$, bootstrap LODA}
  \label{synthetic-table-6}
  \vskip 0.15in
  \centering
  \begin{tabular}{lrrccc|ccc}
    \toprule
        & & \multicolumn{4}{c}{Basic CDF}&\multicolumn{3}{c}{Iso CDF}\\
      &   & \multicolumn{2}{c}{Recall} & \multicolumn{2}{c}{False Positive Rate}&\multicolumn{1}{c}{Recall} & \multicolumn{2}{c}{False Positive Rate} \\
    \cmidrule{2-9}
           $\alpha$   & $n$ & $n^*$ &Recall$\pm$CI&FPR$\pm$CI& Oracle& Recall$\pm$CI &FPR$\pm$CI & Oracle\\
    \midrule
0.01	&	100	&	188760	&	0.719$\pm$0.048	&	0.095$\pm$0.039	&	0.135	&	0.972$\pm$0.017	&	0.569$\pm$0.073	&	0.135	\\
	&	500	&	1065490	&	0.872$\pm$0.022	&	0.083$\pm$0.038	&	0.048	&	0.974$\pm$0.013	&	0.491$\pm$0.079	&	0.048	\\
	&	1000	&	1891769	&	0.899$\pm$0.016	&	0.060$\pm$0.031	&	0.041	&	0.980$\pm$0.009	&	0.511$\pm$0.078	&	0.041	\\
	&	5000	&	5874989	&	0.942$\pm$0.010	&	0.067$\pm$0.030	&	0.034	&	0.988$\pm$0.005	&	0.447$\pm$0.078	&	0.034	\\
	&	10000	&	8907859	&	0.954$\pm$0.008	&	0.069$\pm$0.027	&	0.033	&      0.990$\pm$0.005	&	0.444$\pm$0.075	&	0.033	\\
\bottomrule
  \end{tabular}
 \end{center}
\end{table*}

\begin{table*}
\begin{center}
  \caption{$n^*$, recall (i.e. alien detection rate) and false positive rate from experiments using 9-dimensional normal data, $95\%$, bootstrap LODA}
  \label{synthetic-table-5}
  \vskip 0.15in
  \centering
  \begin{tabular}{lrrccc|ccc}
    \toprule
        & & \multicolumn{4}{c}{Basic CDF}&\multicolumn{3}{c}{Iso CDF}\\
      &   & \multicolumn{2}{c}{Recall} & \multicolumn{2}{c}{False Positive Rate}&\multicolumn{1}{c}{Recall} & \multicolumn{2}{c}{False Positive Rate} \\
    \cmidrule{2-9}
           $\alpha$   & $n$ & $n^*$ &Recall$\pm$CI&FPR$\pm$CI& Oracle& Recall$\pm$CI &FPR$\pm$CI & Oracle\\
    \midrule
0.01	&	100	&	206671	&	0.719$\pm$0.048	&	0.095$\pm$0.039	&	0.077	&	0.972$\pm$0.017	&	0.566$\pm$0.073	&	0.077	\\
	&	500	&	1330996	&	0.872$\pm$0.022	&	0.083$\pm$0.038	&	0.020	&	0.974$\pm$0.013	&	0.487$\pm$0.078	&	0.020	\\
	&	1000	&	2559525	&	0.899$\pm$0.016	&	0.060$\pm$0.031	&	0.015	&	0.978$\pm$0.010	&	0.505$\pm$0.079	&	0.015	\\
	&	5000	&	7369343	&	0.933$\pm$0.010	&	0.052$\pm$0.023	&	0.011	&	0.985$\pm$0.006	&	0.442$\pm$0.078	&	0.011	\\
	&	10000	&	8307313	&	0.943$\pm$0.009	&	0.059$\pm$0.025	&	0.010	&	0.987$\pm$0.006	&	0.413$\pm$0.074	&	0.010	\\
\bottomrule
  \end{tabular}
 \end{center}
\end{table*}

\begin{table*}
\begin{center}
  \caption{$n^*$, recall (i.e. alien detection rate) and false positive rate from experiments using 9-dimensional normal data, $90\%$, bootstrap LODA}
  \label{synthetic-table-4}
  \vskip 0.15in
  \centering
  \begin{tabular}{lrrccc|ccc}
    \toprule
        & & \multicolumn{4}{c}{Basic CDF}&\multicolumn{3}{c}{Iso CDF}\\
      &   & \multicolumn{2}{c}{Recall} & \multicolumn{2}{c}{False Positive Rate}&\multicolumn{1}{c}{Recall} & \multicolumn{2}{c}{False Positive Rate} \\
    \cmidrule{2-9}
           $\alpha$   & $n$ & $n^*$ &Recall$\pm$CI&FPR$\pm$CI& Oracle& Recall$\pm$CI &FPR$\pm$CI & Oracle\\
    \midrule
0.01	&	100	&	242739	&	0.719$\pm$0.048	&	0.095$\pm$0.039	&	0.045	&	0.972$\pm$0.017	&	0.564$\pm$0.073	&	0.045\\
	&	500	&	2059638	&	0.872$\pm$0.022	&	0.082$\pm$0.038	&	0.008	&	0.969$\pm$0.015	&	0.484$\pm$0.079	&	0.008\\
	&	1000	&	3398545	&	0.863$\pm$0.017	&	0.033$\pm$0.020	&	0.006	&	0.973$\pm$0.012	&	0.487$\pm$0.079	&	0.006\\
	&	5000	&	9739214	&	0.900$\pm$0.013	&	0.032$\pm$0.019	&	0.003	&	0.981$\pm$0.007	&	0.411$\pm$0.076	&	0.003\\
	&	10000	&	11575949	&	0.911$\pm$0.012	&	0.030$\pm$0.016	&	0.003	&	0.980$\pm$0.009	&	0.359$\pm$0.072	&	0.003\\
\bottomrule
  \end{tabular}
 \end{center}
\end{table*}

\begin{table*}
\begin{center}
  \caption{1st quartile, median, 3rd quartile of false positive rate from experiments using 9-dimensional normal data, $95\%$, Iforest}
  \label{synthetic-FPR}
  \vskip 0.15in
  \centering
  \begin{tabular}{lrcccc}
    \toprule
        & & \multicolumn{4}{c}{Basic CDF}\\
      &    & \multicolumn{4}{c}{False Positive Rate}\\
    \cmidrule{2-6}
           $\alpha$   & $n$ &1st quartile &median   & 3rd quartile& Oracle(median)\\
    \midrule
0.01	&	100	&	0.004	&	0.006	&	0.014	&	0.051	\\
	&	500	&	0.002	&	0.004	&	0.015	&	0.015	\\
	&	1000	&	0.002	&	0.004	&	0.010	&	0.012	\\
	&	5000	&	0.002	&	0.004	&	0.013	&	0.014	\\
	&	10000	&	0.003	&	0.006	&	0.014	&	0.014	\\
  \cdashline{2-6}
0.05	&	100	&	0.006	&	0.014	&	0.043	&	0.050	\\
	&	500	&	0.004	&	0.008	&	0.023	&	0.015	\\
	&	1000	&	0.004	&	0.008	&	0.015	&	0.012	\\
	&	5000	&	0.006	&	0.010	&	0.020	&	0.014	\\
	&	10000	&	0.008	&	0.011	&	0.019	&	0.014	\\
   \cdashline{2-6}
0.1	&	100	&	0.015	&	0.032	&	0.094	&	0.049	\\
	&	500	&	0.006	&	0.012	&	0.021	&	0.015	\\
	&	1000	&	0.005	&	0.009	&	0.014	&	0.012	\\
	&	5000	&	0.009	&	0.013	&	0.020	&	0.014	\\
	&	10000	&	0.011	&	0.014	&	0.017	&	0.014	\\
    \cdashline{2-6}
0.2	&	100	&	0.025	&	0.043	&	0.105	&	0.049	\\
	&	500	&	0.010	&	0.018	&	0.031	&	0.015	\\
	&	1000	&	0.008	&	0.011	&	0.018	&	0.012	\\
	&	5000	&	0.010	&	0.013	&	0.015	&	0.014	\\
	&	10000	&	0.012	&	0.013	&	0.015	&	0.014	\\
      \cdashline{2-6}
0.5	&	100	&	0.040	&	0.058	&	0.090	&	0.051	\\
	&	500	&	0.012	&	0.016	&	0.021	&	0.015	\\
	&	1000	&	0.011	&	0.012	&	0.016	&	0.012	\\
	&	5000	&	0.013	&	0.014	&	0.016	&	0.014	\\
	&	10000	&	0.013	&	0.014	&	0.015	&	0.014	\\
    \bottomrule
  \end{tabular}
 \end{center}
\end{table*}



\section{Experimental Results from UCI and Image Datasets}
In this section we include results of performance on UCI benchmarks, MNIST and Tiny Imagenet and Tables 8-22 illustrate the results.  The experimental protocol is similar to synthetic datasets and two state of the art anomaly detectors Isolation forest, LODA are applied. For Isolation forest we train Forest with 1000 trees on nominal dataset and use out of bag estimates of this dataset to estimate the nominal datasets anomaly score distribution. For LODA we build 1000 projections and similar to Isolation forest we get anomaly score for each point in nominal dataset using the projections that didn't use this point.Tables 11-16 illustrate the results of LODA for 6  different datasets for varying values of $\eta$ and report the observed recall, False positive rate averaged over 100 runs of each experiment. Tables 17-22 report the results for Isolation Forest and it can be observed the performance of both LODA,Isolation Forest are similar.

For Image datasets we follow the same protocol as UCI for MNIST and apply Isolation Forest on the input image but for Tiny Imagenet the anomaly scores are obtained differently. We first train a Wide Residual Network (40-2) classifier on the 200 nominal classes of Tiny Imagenet and apply baseline method \cite{hendrycks17baseline} on validation data to get the nominal dataset distribution and later apply the same method on the mixture dataset which will have $\alpha$ proportion of aliens which are basically from 800 held out classes.Tables 8-10 illustrate the results for these datasets for target recall of 98\%,95\% and 90\%.

\begin{table*}
\begin{center}
\centering
\caption{Recall (i.e. alien detection rate) \& False Positive Rate for Image Datasets,98\%}
\label{my-label-Image-1}
\vskip 0.15in
\begin{tabular}{lcc|ll|ll}
\toprule
         & & &  \multicolumn{2}{c}{Basic CDF}&\multicolumn{2}{c}{Iso CDF}\\
 		& &  &Recall&False Positive Rate  &Recall &False Positive Rate\\
		\cmidrule{2-7}
Dataset &$\alpha$ & $\hat\alpha$  &recall$\pm$CI &FPR$\pm$CI   &recall$\pm$CI &FPR$\pm$CI\\
\midrule
Tiny Image Net & 0.100 & 0.100 & 0.926 $\pm$ 0.014 & 0.677 $\pm$ 0.030 & 0.944 $\pm$ 0.013 & 0.746 $\pm$ 0.034 \\
n=10000 & 0.100 & 0.104 & 0.933 $\pm$ 0.013 & 0.695 $\pm$ 0.030 & 0.952 $\pm$ 0.012 & 0.766 $\pm$ 0.033 \\
 & 0.100 & 0.108 & 0.941 $\pm$ 0.012 & 0.715 $\pm$ 0.029 & 0.959 $\pm$ 0.011 & 0.786 $\pm$ 0.031 \\
 \cdashline{2-7}
& 0.200 & 0.200 & 0.965 $\pm$ 0.005 & 0.738 $\pm$ 0.018 & 0.972 $\pm$ 0.005 & 0.774 $\pm$ 0.021 \\
 & 0.200 & 0.204 & 0.970 $\pm$ 0.004 & 0.761 $\pm$ 0.018 & 0.977 $\pm$ 0.004 & 0.798 $\pm$ 0.019 \\
\bottomrule
  \end{tabular}
 \end{center}
\end{table*}

\begin{table*}
\begin{center}
\centering
\caption{Recall (i.e. alien detection rate) \& False Positive Rate for Image Datasets,95\%}
\label{my-label-Image-2}
\vskip 0.15in
\begin{tabular}{lcc|ll|ll}
\toprule
         & & &  \multicolumn{2}{c}{Basic CDF}&\multicolumn{2}{c}{Iso CDF}\\
     
 		& &  &Recall&False Positive Rate  &Recall &False Positive Rate\\
		\cmidrule{2-7}
Dataset &$\alpha$ & $\hat\alpha$  &recall$\pm$CI &FPR$\pm$CI   &recall$\pm$CI &FPR$\pm$CI\\
\midrule
Tiny Image Net & 0.100 & 0.100 & 0.902 $\pm$ 0.014 & 0.620 $\pm$ 0.025 & 0.924 $\pm$ 0.014 & 0.686 $\pm$ 0.032 \\
n=10000 & 0.100 & 0.104 & 0.912 $\pm$ 0.014 & 0.639 $\pm$ 0.026 & 0.930 $\pm$ 0.014 & 0.697 $\pm$ 0.031 \\
 & 0.100 & 0.108 & 0.923 $\pm$ 0.013 & 0.660 $\pm$ 0.026 & 0.939 $\pm$ 0.012 & 0.716 $\pm$ 0.031 \\
 \cdashline{2-7}
& 0.200 & 0.200 & 0.942 $\pm$ 0.006 & 0.667 $\pm$ 0.016 & 0.948 $\pm$ 0.006 & 0.682 $\pm$ 0.016 \\
 & 0.200 & 0.204 & 0.949 $\pm$ 0.005 & 0.683 $\pm$ 0.014 & 0.954 $\pm$ 0.005 & 0.700 $\pm$ 0.015 \\
\bottomrule
  \end{tabular}
 \end{center}
\end{table*}

\begin{table*}
\begin{center}
\centering
\caption{Recall (i.e. alien detection rate) \& False Positive Rate for Image Datasets,90\%}
\label{my-label-Image-3}
\vskip 0.15in
\begin{tabular}{lcc|ll|ll}
\toprule
         & & &  \multicolumn{2}{c}{Basic CDF}&\multicolumn{2}{c}{Iso CDF}\\
     
 		& &  &Recall&False Positive Rate  &Recall &False Positive Rate\\
		\cmidrule{2-7}
Dataset &$\alpha$ & $\hat\alpha$  &recall$\pm$CI &FPR$\pm$CI   &recall$\pm$CI &FPR$\pm$CI\\
\midrule
Tiny Image Net & 0.100 & 0.100 & 0.862 $\pm$ 0.015 & 0.545 $\pm$ 0.021 & 0.883 $\pm$ 0.016 & 0.590 $\pm$ 0.027 \\
n=10000 & 0.100 & 0.104 & 0.873 $\pm$ 0.015 & 0.562 $\pm$ 0.021 & 0.892 $\pm$ 0.015 & 0.602 $\pm$ 0.026 \\
 & 0.100 & 0.108 & 0.885 $\pm$ 0.014 & 0.579 $\pm$ 0.021 & 0.902 $\pm$ 0.014 & 0.617 $\pm$ 0.025 \\
  \cdashline{2-7}
 & 0.200 & 0.200 & 0.894 $\pm$ 0.007 & 0.578 $\pm$ 0.011 & 0.898 $\pm$ 0.007 & 0.584 $\pm$ 0.011 \\
 & 0.200 & 0.204 & 0.904 $\pm$ 0.007 & 0.593 $\pm$ 0.010 & 0.909 $\pm$ 0.006 & 0.600 $\pm$ 0.011 \\
\bottomrule
  \end{tabular}
 \end{center}
\end{table*}

\begin{table*}
\begin{center}
\centering
\caption{Recall \& False Positive Rate for Landsat Dataset using LODA for varying $q$ (target recall $1-q$)}
\label{my-label-4}
\vskip 0.15in
\begin{tabular}{lccc|ll|ll}
\toprule
         & & & & \multicolumn{2}{c}{Basic CDF}&\multicolumn{2}{c}{Iso CDF}\\
 		& &  & &Recall&False Positive Rate  &Recall &False Positive Rate\\
		\cmidrule{2-8}
Dataset  &$\alpha$ & $\hat\alpha$ &$q$ &recall$\pm$CI &FPR$\pm$CI   &recall$\pm$CI &FPR$\pm$CI\\
\midrule
Landsat & 0.100 & 0.100 & 0.020 & 0.937 $\pm$ 0.024 & 0.162 $\pm$ 0.047 & 0.960 $\pm$ 0.024 & 0.495 $\pm$ 0.068 \\
n=1532 & 0.100 & 0.104 & 0.020 & 0.949 $\pm$ 0.024 & 0.203 $\pm$ 0.048 & 0.967 $\pm$ 0.024 & 0.543 $\pm$ 0.064 \\
 & 0.100 & 0.108 & 0.020 & 0.960 $\pm$ 0.024 & 0.255 $\pm$ 0.050 & 0.972 $\pm$ 0.024 & 0.583 $\pm$ 0.062 \\
\cdashline{2-8}
& 0.200 & 0.200 & 0.020 & 0.965 $\pm$ 0.005 & 0.128 $\pm$ 0.033 & 0.983 $\pm$ 0.005 & 0.404 $\pm$ 0.062 \\
 & 0.200 & 0.204 & 0.020 & 0.980 $\pm$ 0.003 & 0.204 $\pm$ 0.040 & 0.991 $\pm$ 0.003 & 0.478 $\pm$ 0.061 \\
\bottomrule
  \end{tabular}
 \end{center}
\end{table*}
   \begin{table*}
\begin{center}
\centering
\caption{Recall \& False Positive Rate for page.blocks Dataset using LODA for varying $q$ (target recall $1-q$)}
\label{my-label-5}
\vskip 0.15in
\begin{tabular}{lccc|ll|ll}
\toprule
         & & & & \multicolumn{2}{c}{Basic CDF}&\multicolumn{2}{c}{Iso CDF}\\
 		& &  & &Recall&False Positive Rate  &Recall &False Positive Rate\\
		\cmidrule{2-8}
Dataset  &$\alpha$ & $\hat\alpha$ &$q$ &recall$\pm$CI &FPR$\pm$CI   &recall$\pm$CI &FPR$\pm$CI\\
\midrule
pageblocks & 0.100 & 0.100 & 0.020 & 0.963 $\pm$ 0.021 & 0.254 $\pm$ 0.084 & 0.983 $\pm$ 0.013 & 0.555 $\pm$ 0.201 \\
n=4912 & 0.100 & 0.104 & 0.020 & 0.969 $\pm$ 0.017 & 0.315 $\pm$ 0.138 & 0.991 $\pm$ 0.006 & 0.624 $\pm$ 0.181 \\
 & 0.100 & 0.108 & 0.020 & 0.976 $\pm$ 0.013 & 0.357 $\pm$ 0.137 & 0.995 $\pm$ 0.004 & 0.712 $\pm$ 0.159 \\
 \cdashline{2-8}
 & 0.200 & 0.200 & 0.020 & 0.966 $\pm$ 0.005 & 0.287 $\pm$ 0.031 & 0.978 $\pm$ 0.005 & 0.452 $\pm$ 0.053 \\
 & 0.200 & 0.204 & 0.020 & 0.978 $\pm$ 0.004 & 0.367 $\pm$ 0.038 & 0.987 $\pm$ 0.003 & 0.529 $\pm$ 0.050 \\
\bottomrule
  \end{tabular}
 \end{center}
\end{table*}

   \begin{table*}
\begin{center}
\centering
\caption{Recall \& False Positive Rate for Optical.digits Dataset using LODA for varying $q$ (target recall $1-q$)}
\label{my-label-6}
\vskip 0.15in
\begin{tabular}{lccc|ll|ll}
\toprule
         & & & & \multicolumn{2}{c}{Basic CDF}&\multicolumn{2}{c}{Iso CDF}\\
 		& &  & &Recall&False Positive Rate  &Recall &False Positive Rate\\
		\cmidrule{2-8}
Dataset  &$\alpha$ & $\hat\alpha$ &$q$ &recall$\pm$CI &FPR$\pm$CI   &recall$\pm$CI &FPR$\pm$CI\\
\midrule
Optical.digits & 0.100 & 0.100 & 0.020 & 0.898 $\pm$ 0.016 & 0.167 $\pm$ 0.035 & 0.947 $\pm$ 0.016 & 0.502 $\pm$ 0.070 \\
n=568 & 0.100 & 0.104 & 0.020 & 0.906 $\pm$ 0.013 & 0.186 $\pm$ 0.036 & 0.950 $\pm$ 0.013 & 0.519 $\pm$ 0.066 \\
 & 0.100 & 0.108 & 0.020 & 0.917 $\pm$ 0.013 & 0.214 $\pm$ 0.036 & 0.957 $\pm$ 0.012 & 0.549 $\pm$ 0.065 \\
 \cdashline{2-8}
 & 0.200 & 0.200 & 0.020 & 0.938 $\pm$ 0.011 & 0.177 $\pm$ 0.033 & 0.964 $\pm$ 0.010 & 0.434 $\pm$ 0.065 \\
 & 0.200 & 0.204 & 0.020 & 0.953 $\pm$ 0.008 & 0.220 $\pm$ 0.039 & 0.973 $\pm$ 0.008 & 0.466 $\pm$ 0.063 \\
\bottomrule
  \end{tabular}
 \end{center}
\end{table*}

   \begin{table*}
\begin{center}
\centering
\caption{Recall \& False Positive Rate for Letter Recognition Dataset using LODA for varying $q$ (target recall $1-q$)}
\label{my-label-7}
\vskip 0.15in
\begin{tabular}{lccc|ll|ll}
\toprule
         & & & & \multicolumn{2}{c}{Basic CDF}&\multicolumn{2}{c}{Iso CDF}\\
 		& &  & &Recall&False Positive Rate  &Recall &False Positive Rate\\
		\cmidrule{2-8}
Dataset  &$\alpha$ & $\hat\alpha$ &$q$ &recall$\pm$CI &FPR$\pm$CI   &recall$\pm$CI &FPR$\pm$CI\\
\midrule
Letter recognition & 0.100 & 0.100 & 0.020 & 0.897 $\pm$ 0.018 & 0.232 $\pm$ 0.044 & 0.953 $\pm$ 0.016 & 0.596 $\pm$ 0.069 \\
n=788 & 0.100 & 0.104 & 0.020 & 0.918 $\pm$ 0.013 & 0.255 $\pm$ 0.045 & 0.967 $\pm$ 0.010 & 0.613 $\pm$ 0.067 \\
 & 0.100 & 0.108 & 0.020 & 0.931 $\pm$ 0.012 & 0.291 $\pm$ 0.047 & 0.975 $\pm$ 0.008 & 0.633 $\pm$ 0.065 \\
 \cdashline{2-8}
 & 0.200 & 0.200 & 0.020 & 0.940 $\pm$ 0.011 & 0.197 $\pm$ 0.032 & 0.964 $\pm$ 0.011 & 0.457 $\pm$ 0.066 \\
 & 0.200 & 0.204 & 0.020 & 0.954 $\pm$ 0.008 & 0.235 $\pm$ 0.035 & 0.974 $\pm$ 0.007 & 0.490 $\pm$ 0.063 \\
\bottomrule
  \end{tabular}
 \end{center}
\end{table*}

   \begin{table*}
\begin{center}
\centering
\caption{Recall \& False Positive Rate for Shuttle Dataset using LODA for varying $q$ (target recall $1-q$)}
\label{my-label-8}
\vskip 0.15in
\begin{tabular}{lccc|ll|ll}
\toprule
         & & & & \multicolumn{2}{c}{Basic CDF}&\multicolumn{2}{c}{Iso CDF}\\
 		& &  & &Recall&False Positive Rate  &Recall &False Positive Rate\\
		\cmidrule{2-8}
Dataset  &$\alpha$ & $\hat\alpha$ &$q$ &recall$\pm$CI &FPR$\pm$CI   &recall$\pm$CI &FPR$\pm$CI\\
\midrule
Shuttle & 0.100 & 0.100 & 0.020 & 0.959 $\pm$ 0.008 & 0.145 $\pm$ 0.031 & 0.973 $\pm$ 0.007 & 0.425 $\pm$ 0.065 \\
n=5000 & 0.100 & 0.104 & 0.020 & 0.971 $\pm$ 0.006 & 0.205 $\pm$ 0.038 & 0.984 $\pm$ 0.006 & 0.490 $\pm$ 0.065 \\
 & 0.100 & 0.108 & 0.020 & 0.984 $\pm$ 0.005 & 0.287 $\pm$ 0.048 & 0.991 $\pm$ 0.003 & 0.550 $\pm$ 0.064 \\
\cdashline{2-8}
& 0.200 & 0.200 & 0.020 & 0.975 $\pm$ 0.004 & 0.100 $\pm$ 0.016 & 0.982 $\pm$ 0.004 & 0.317 $\pm$ 0.058 \\
 & 0.200 & 0.204 & 0.020 & 0.990 $\pm$ 0.002 & 0.195 $\pm$ 0.033 & 0.994 $\pm$ 0.002 & 0.451 $\pm$ 0.058 \\
\bottomrule
  \end{tabular}
 \end{center}
\end{table*}

   \begin{table*}
\begin{center}
\centering
\caption{Recall \& False Positive Rate for Covertype Dataset using LODA for varying $q$ (target recall $1-q$)}
\label{my-label-9}
\vskip 0.15in
\begin{tabular}{lccc|ll|ll}
\toprule
         & & & & \multicolumn{2}{c}{Basic CDF}&\multicolumn{2}{c}{Iso CDF}\\
 		& &  & &Recall&False Positive Rate  &Recall &False Positive Rate\\
		\cmidrule{2-8}
Dataset  &$\alpha$ & $\hat\alpha$ &$q$ &recall$\pm$CI &FPR$\pm$CI   &recall$\pm$CI &FPR$\pm$CI\\
\midrule
Covertype & 0.100 & 0.100 & 0.020 & 0.979 $\pm$ 0.002 & 0.010 $\pm$ 0.003 & 0.989 $\pm$ 0.002 & 0.373 $\pm$ 0.072 \\
n=13624 & 0.100 & 0.104 & 0.020 & 0.995 $\pm$ 0.001 & 0.098 $\pm$ 0.025 & 0.998 $\pm$ 0.001 & 0.470 $\pm$ 0.066 \\
 & 0.100 & 0.108 & 0.020 & 0.999 $\pm$ 0.000 & 0.292 $\pm$ 0.045 & 0.999 $\pm$ 0.000 & 0.586 $\pm$ 0.059 \\
\cdashline{2-8}
& 0.200 & 0.200 & 0.020 & 0.980 $\pm$ 0.001 & 0.007 $\pm$ 0.002 & 0.987 $\pm$ 0.002 & 0.220 $\pm$ 0.056 \\
 & 0.200 & 0.204 & 0.020 & 0.998 $\pm$ 0.000 & 0.142 $\pm$ 0.026 & 0.999 $\pm$ 0.000 & 0.419 $\pm$ 0.057 \\
\bottomrule
  \end{tabular}
 \end{center}
\end{table*}
   \begin{table*}
\begin{center}
\centering
\caption{Recall \& False Positive Rate for Landsat Dataset using Iforest for varying $q$ (target recall $1-q$)}
\label{my-label-10}
\vskip 0.15in
\begin{tabular}{lccc|ll|ll}
\toprule
         & & & & \multicolumn{2}{c}{Basic CDF}&\multicolumn{2}{c}{Iso CDF}\\
    & &  & &Recall&False Positive Rate  &Recall &False Positive Rate\\
    \cmidrule{2-8}
Dataset  &$\alpha$ & $\hat\alpha$ &$q$ &recall$\pm$CI &FPR$\pm$CI   &recall$\pm$CI &FPR$\pm$CI\\
\midrule
Landsat & 0.100 & 0.100 & 0.020 & 0.941 $\pm$ 0.024 & 0.164 $\pm$ 0.045 & 0.964 $\pm$ 0.025 & 0.503 $\pm$ 0.066 \\
n=1532 & 0.100 & 0.104 & 0.020 & 0.950 $\pm$ 0.024 & 0.197 $\pm$ 0.048 & 0.968 $\pm$ 0.024 & 0.545 $\pm$ 0.065 \\
 & 0.100 & 0.108 & 0.020 & 0.960 $\pm$ 0.024 & 0.254 $\pm$ 0.051 & 0.971 $\pm$ 0.024 & 0.584 $\pm$ 0.063 \\
 \cdashline{2-8}
& 0.200 & 0.200 & 0.020 & 0.965 $\pm$ 0.005 & 0.130 $\pm$ 0.033 & 0.982 $\pm$ 0.004 & 0.402 $\pm$ 0.063 \\
 & 0.200 & 0.204 & 0.020 & 0.979 $\pm$ 0.003 & 0.199 $\pm$ 0.040 & 0.991 $\pm$ 0.003 & 0.477 $\pm$ 0.060 \\
\bottomrule
  \end{tabular}
 \end{center}
\end{table*}

   \begin{table*}
\begin{center}
\centering
\caption{Recall \& False Positive Rate for page.blocks Dataset using Iforest for varying $q$ (target recall $1-q$)}
\label{my-label-11}
\vskip 0.15in
\begin{tabular}{lccc|ll|ll}
\toprule
         & & & & \multicolumn{2}{c}{Basic CDF}&\multicolumn{2}{c}{Iso CDF}\\
 		& &  & &Recall&False Positive Rate  &Recall &False Positive Rate\\
		\cmidrule{2-8}
Dataset  &$\alpha$ & $\hat\alpha$ &$q$ &recall$\pm$CI &FPR$\pm$CI   &recall$\pm$CI &FPR$\pm$CI\\
\midrule
page.blocks & 0.100 & 0.100 & 0.020 & 0.951 $\pm$ 0.029 & 0.269 $\pm$ 0.106 & 0.975 $\pm$ 0.017 & 0.511 $\pm$ 0.220 \\
n=4912 & 0.100 & 0.104 & 0.020 & 0.968 $\pm$ 0.018 & 0.314 $\pm$ 0.130 & 0.991 $\pm$ 0.007 & 0.641 $\pm$ 0.184 \\
 & 0.100 & 0.108 & 0.020 & 0.976 $\pm$ 0.013 & 0.366 $\pm$ 0.138 & 0.994 $\pm$ 0.005 & 0.692 $\pm$ 0.175 \\
\midrule
& 0.200 & 0.200 & 0.020 & 0.965 $\pm$ 0.005 & 0.283 $\pm$ 0.030 & 0.976 $\pm$ 0.005 & 0.443 $\pm$ 0.052 \\
 & 0.200 & 0.204 & 0.020 & 0.979 $\pm$ 0.004 & 0.366 $\pm$ 0.037 & 0.986 $\pm$ 0.004 & 0.527 $\pm$ 0.051 \\
\bottomrule
  \end{tabular}
 \end{center}
\end{table*}

   \begin{table*}
\begin{center}
\centering
\caption{Recall \& False Positive Rate for Optical.digits Dataset using Iforest for varying $q$ (target recall $1-q$)}
\label{my-label-12}
\vskip 0.15in
\begin{tabular}{lccc|ll|ll}
\toprule
         & & & & \multicolumn{2}{c}{Basic CDF}&\multicolumn{2}{c}{Iso CDF}\\
 		& &  & &Recall&False Positive Rate  &Recall &False Positive Rate\\
		\cmidrule{2-8}
Dataset  &$\alpha$ & $\hat\alpha$ &$q$ &recall$\pm$CI &FPR$\pm$CI   &recall$\pm$CI &FPR$\pm$CI\\
\midrule
Optical.digits & 0.100 & 0.100 & 0.020 & 0.891 $\pm$ 0.018 & 0.172 $\pm$ 0.036 & 0.940 $\pm$ 0.017 & 0.504 $\pm$ 0.068 \\
n=568 & 0.100 & 0.104 & 0.020 & 0.904 $\pm$ 0.013 & 0.191 $\pm$ 0.036 & 0.952 $\pm$ 0.012 & 0.521 $\pm$ 0.066 \\
 & 0.100 & 0.108 & 0.020 & 0.917 $\pm$ 0.012 & 0.214 $\pm$ 0.037 & 0.961 $\pm$ 0.011 & 0.548 $\pm$ 0.065 \\
\cdashline{2-8}
& 0.200 & 0.200 & 0.020 & 0.941 $\pm$ 0.010 & 0.178 $\pm$ 0.032 & 0.964 $\pm$ 0.010 & 0.424 $\pm$ 0.065 \\
 & 0.200 & 0.204 & 0.020 & 0.951 $\pm$ 0.008 & 0.218 $\pm$ 0.037 & 0.974 $\pm$ 0.008 & 0.465 $\pm$ 0.064 \\
\bottomrule
  \end{tabular}
 \end{center}
\end{table*}

   \begin{table*}
\begin{center}
\centering
\caption{Recall \& False Positive Rate for Letter Recognition Dataset using Iforest for varying $q$ (target recall $1-q$)}
\label{my-label-13}
\vskip 0.15in
\begin{tabular}{lccc|ll|ll}
\toprule
         & & & & \multicolumn{2}{c}{Basic CDF}&\multicolumn{2}{c}{Iso CDF}\\
 		& &  & &Recall&False Positive Rate  &Recall &False Positive Rate\\
		\cmidrule{2-8}
Dataset  &$\alpha$ & $\hat\alpha$ &$q$ &recall$\pm$CI &FPR$\pm$CI   &recall$\pm$CI &FPR$\pm$CI\\
\midrule
Letter recog & 0.100 & 0.100 & 0.020 & 0.911 $\pm$ 0.016 & 0.234 $\pm$ 0.044 & 0.960 $\pm$ 0.013 & 0.588 $\pm$ 0.070 \\
n=788 & 0.100 & 0.104 & 0.020 & 0.919 $\pm$ 0.013 & 0.262 $\pm$ 0.046 & 0.965 $\pm$ 0.011 & 0.603 $\pm$ 0.067 \\
 & 0.100 & 0.108 & 0.020 & 0.931 $\pm$ 0.012 & 0.297 $\pm$ 0.048 & 0.973 $\pm$ 0.009 & 0.627 $\pm$ 0.065 \\
\cdashline{2-8}
& 0.200 & 0.200 & 0.020 & 0.936 $\pm$ 0.011 & 0.203 $\pm$ 0.034 & 0.961 $\pm$ 0.011 & 0.444 $\pm$ 0.065 \\
 & 0.200 & 0.204 & 0.020 & 0.954 $\pm$ 0.008 & 0.236 $\pm$ 0.036 & 0.975 $\pm$ 0.008 & 0.487 $\pm$ 0.064 \\
\bottomrule
  \end{tabular}
 \end{center}
\end{table*}

   \begin{table*}
\begin{center}
\centering
\caption{Recall \& False Positive Rate for Shuttle Dataset using Iforest for varying $q$ (target recall $1-q$)}
\label{my-label-14}
\vskip 0.15in
\begin{tabular}{lccc|ll|ll}
\toprule
         & & & & \multicolumn{2}{c}{Basic CDF}&\multicolumn{2}{c}{Iso CDF}\\
 		& &  & &Recall&False Positive Rate  &Recall &False Positive Rate\\
		\cmidrule{2-8}
Dataset  &$\alpha$ & $\hat\alpha$ &$q$ &recall$\pm$CI &FPR$\pm$CI   &recall$\pm$CI &FPR$\pm$CI\\
\midrule
Shuttle & 0.100 & 0.100 & 0.020 & 0.959 $\pm$ 0.008 & 0.147 $\pm$ 0.031 & 0.974 $\pm$ 0.007 & 0.419 $\pm$ 0.066 \\
n=5000 & 0.100 & 0.104 & 0.020 & 0.972 $\pm$ 0.006 & 0.206 $\pm$ 0.039 & 0.985 $\pm$ 0.005 & 0.490 $\pm$ 0.065 \\
 & 0.100 & 0.108 & 0.020 & 0.984 $\pm$ 0.004 & 0.287 $\pm$ 0.047 & 0.991 $\pm$ 0.003 & 0.545 $\pm$ 0.064 \\
\cdashline{2-8}
& 0.200 & 0.200 & 0.020 & 0.975 $\pm$ 0.004 & 0.103 $\pm$ 0.016 & 0.983 $\pm$ 0.004 & 0.326 $\pm$ 0.058 \\
 & 0.200 & 0.204 & 0.020 & 0.990 $\pm$ 0.002 & 0.192 $\pm$ 0.033 & 0.994 $\pm$ 0.002 & 0.457 $\pm$ 0.058 \\
\bottomrule
  \end{tabular}
 \end{center}
\end{table*}

   \begin{table*}
\begin{center}
\centering
\caption{Recall \& False Positive Rate for Covertype Dataset using Iforest for varying $q$ (target recall $1-q$)}
\label{my-label-15}
\vskip 0.15in
\begin{tabular}{lccc|ll|ll}
\toprule
         & & & & \multicolumn{2}{c}{Basic CDF}&\multicolumn{2}{c}{Iso CDF}\\
 		& &  & &Recall&False Positive Rate  &Recall &False Positive Rate\\
		\cmidrule{2-8}
Dataset  &$\alpha$ & $\hat\alpha$ &$q$ &recall$\pm$CI &FPR$\pm$CI   &recall$\pm$CI &FPR$\pm$CI\\
\midrule
Covertype & 0.100 & 0.100 & 0.020 & 0.979 $\pm$ 0.002 & 0.012 $\pm$ 0.007 & 0.989 $\pm$ 0.002 & 0.359 $\pm$ 0.072 \\
n=13624 & 0.100 & 0.104 & 0.020 & 0.995 $\pm$ 0.001 & 0.095 $\pm$ 0.025 & 0.998 $\pm$ 0.001 & 0.479 $\pm$ 0.067 \\
 & 0.100 & 0.108 & 0.020 & 0.999 $\pm$ 0.000 & 0.292 $\pm$ 0.045 & 0.999 $\pm$ 0.000 & 0.586 $\pm$ 0.059 \\
\cdashline{2-8}
& 0.200 & 0.200 & 0.020 & 0.980 $\pm$ 0.001 & 0.006 $\pm$ 0.002 & 0.986 $\pm$ 0.002 & 0.211 $\pm$ 0.055 \\
 & 0.200 & 0.204 & 0.020 & 0.998 $\pm$ 0.000 & 0.143 $\pm$ 0.027 & 0.999 $\pm$ 0.000 & 0.420 $\pm$ 0.058 \\
\bottomrule
  \end{tabular}
 \end{center}
\end{table*}

\end{document}